# 面向具身操作的视觉-语言-动作模型综述

李浩然[1,3] 陈宇辉[1] 崔文博[1] 刘卫恒[1] 刘 锴[1]
周明才[1,2] 张正涛[1,2] 赵冬斌[1,3,✉]

**摘 要** 具身智能系统通过智能体与环境不断交互，从而提升智能体能力，受到了学术界和产业界的广泛关注。视觉-语言-动作模型作为一种受到大模型发展启发的机器人通用控制模型，提高了具身智能系统中智能体与环境交互的能力，大大扩展了具身智能机器人的应用场景。本文对具身操作中的视觉-语言-动作模型进行了综述，首先，详细介绍了视觉-语言-动作模型的发展历程，然后，对视觉-语言-动作模型架构、训练数据、预训练方法、后训练方法和模型评估 5 个方面的研究现状进行了详细地分析，最后，针对视觉-语言-动作模型发展过程和机器人操作落地应用中面临的挑战和未来可能的发展方向进行了总结。

## Survey of Vision-Language-Action Models for Embodied Manipulation

LI Hao-Ran[1,3] CHEN Yu-Hui[1] CUI Wen-Bo[1] LIU Wei-Heng[1] LIU Kai[1]
ZHOU Ming-Cai[1,2] ZHANG Zheng-Tao[1,2] ZHAO Dong-Bin[1,3,✉]

**Abstract** Embodied intelligence systems, which enhance agent capabilities through continuous environment interactions, have garnered significant attention from both academia and industry. Vision-Language-Action models, inspired by advancements in large foundation models, serve as universal robotic control frameworks that substantially improve agent-environment interaction capabilities in embodied intelligence systems. This expansion has broadened application scenarios for embodied AI robots. This survey comprehensively reviews VLA models for embodied manipulation. Firstly, it chronicles the developmental trajectory of VLA architectures. Subsequently, we conduct a detailed analysis of current research across 5 critical dimensions: VLA model structures, training datasets, pre-training methods, post-training methods, and model evaluation. Finally, we synthesize key challenges in VLA development and real-world deployment, while outlining promising future research directions.

近年来，具身智能受到了学术界和产业界的广泛关注。相比于传统的互联网智能或离身智能从数据中获取智能，具身智能系统通过控制“本体”与环境交互，从而获得智能。作为具身智能“本体”的典型代表之一，操作机器人通过控制机械臂与环境进行交互完成抓放、搬运等任务，在工业生产中广泛应用。传统的机器人系统通常由多个不同的模块组合构成，例如感知模型通过处理传感器数据获取环境状态和操作对象状态，决策模型则根据当前的状态确定操作目标。接收到决策目标后，规划模型根据机械臂与环境状态规划可行路径，最后由控制模型控制机械臂跟踪该路径以完成任务。这种模块化的机器人系统可以从各个模块的发展中受益，并且具有良好的可解释性。但是当操作机器人推广到日常生活中，基于逻辑编排的决策模块和基于搜索或优化的规划与控制模块很难应对开放环境下多样性任务需求以及复杂的交互行为。

随着大语言模型(Large Language Model, LLM)和视觉语言模型 (Vision Language Model, VLM) 的兴起,以 Transformer[6] 结构为核心的基础模型展现

收稿日期 2025-08-25 录用日期 2025-11-06
Manuscript received August 25, 2025; accepted November 06, 2025
国家自然科学基金 (62136008, 62173324) 资助
Supported by National Natural Science Foundation of China (62136008, 62173324)
本文责任编委 XXX
Recommended by Associate Editor XXX

1. 中国科学院自动化研究所 北京 100190 2. 北京中科慧灵机器人技术有限公司 北京 100080 3. 中国科学院大学人工智能学院 北京 101408
1. Institute of Automation, Chinese Academy of Sciences, Beijing 100190 2. Beijing Zhongke Huiling Robot Technology Co., Beijing 100080 3. School of Artificial Intelligence, University of Chinese Academy of Sciences, Beijing, 101408

表 1: 本文与其他 VLA 相关综述的对比
Tab. 1: Comparison with Other VLA-related Surveys.

| 综述 | 发展历程 | 模型结构 | 数据集 | 预训练方法 | 后训练方法 | 模型评估 |
|---|---|---|---|---|---|---|
| Ma et al.[1] | ✓ | ✓ | ✓ | ✓ | ✗ | ✓ |
| Sapkota et al.[2] | ✓ | ✓ | ✗ | ✓ | ✗ | ✗ |
| Zhong et al.[3] | ✓ | ✓ | ✓ | ✗ | ✗ | ✗ |
| Xiang et al.[4] | ✗ | ✓ | ✗ | ✗ | ✓ | ✗ |
| Din et al.[5] | ✓ | ✓ | ✓ | ✗ | ✗ | ✓ |
| 本文 | ✓ | ✓ | ✓ | ✓ | ✓ | ✓ |

出了强大的泛化能力，机器人技术也迎来了新的发展机遇。通过大模型强大的视觉理解能力和自然语言理解能力，机器人在任务规划和开放环境适应性方面取得了显著进步。例如，机器人可以通过 VLM 识别物体并根据语言指令规划路径[7]，从而在一定程度上应对环境的多变性和任务的多样性。也可以让大模型根据当前环境和任务指令，生成相应的代码，从而指导机器人动作[8]。然而，这种模式存在语义理解与物理执行的割裂。大模型主要承担环境理解和规划层功能，无法理解机器人的执行能力。然而大模型规划的动作需依赖预编程的下层控制器，导致机器人任务理解与执行出现脱节，无法实现复杂的动作行为。

相比于传统多模块解耦系统容易受到模块短板效应的影响，视觉模仿学习通过直接建立视觉图像与机器人动作之间的映射关系，从而可以实现更灵活的机器人运动控制。但是早期的方法大部分局限在特定任务或数据上，往往难以适应新任务或多变的环境，限制了机器人在复杂场景中的应用潜力。随着大模型技术的快速发展，LLM 和 VLM 展现出优秀的语义理解和泛化能力，让实现开放环境下的通用机器人策略成为可能。视觉-语言-动作（Vision Language Action, VLA）模型通过结合大模型技术，将视觉感知、语义推理与动作生成深度融合，使机器人能够直接从多模态输入中预测连续控制指令，实现从环境理解到物理执行的闭环耦合。目前，以 VLA 为核心的机器人系统在开放指令抓取、柔性物体操作、双臂协作、以及多机器人协作等领域展现出令人印象深刻的性能，大大提高了人们对于机器人任务的想象力。

从发展历程看，VLA 的发展很大程度上受到了大模型发展的启发。Bai et al.[9]，Wen-Sheng Wang[10] 详细分析了大模型在具身智能系统的感知，规划，决策，数据生成等方面发挥的作用。Ma et al.[1] 对 2024 年之前的 VLA 方法进行了整理，综述内容虽然涵盖了模型的发展历程、模型结构、数据集、训练方法和模型评估等多个方面，但是一方面由于 2024 年之后 VLA 技术迭代非常快，技术路线与之前已经呈现显著的差异性，其所阐述的模型结构、数据和训练方法已不具备代表性。另一方面，该综述只从仿真器测评基准的角度阐述了 VLA 模型评估，很难适应当前的应用需求。Sapkota et al.[2] 较为全面地从 VLA 概念、发展历程、模型结构、训练方法和 VLA 应用等多个角度对 2025 年上半年之前的 VLA 模型提供了全面综述。Zhong et al.[3] 从动作空间的角度分析了目前 VLA 的发展现状，详细阐述了不同动作空间下 VLA 的实现方式，局限性和未来趋势。Xiang et al.[4] 从人类运动学习的角度，对于 VLA 模型的后训练方法进行了梳理与分析。Din et al.[5] 梳理了 VLA 模型结构和发展历程，并对 VLA 训练数据进行了详细的阐述，并且通过成功率和零样本泛化能力对当前 VLA 模型进行了定性评估。与上述工作不同的是，本文从具身智能系统的角度出发，针对环境、本体和进化算法 3 个具身系统的核心要素，从模型结构、训练所使用的数据集、预训练方法、后训练方法以及模型评估 5 个方面，审视目前 VLA 的发展现状，深入剖析 VLA 模型的核心构成，阐述每个部分在具身操作场景下面临的困境和未来的潜在发展方向，希望能够为该领域研究人员提供参考和方向指导。具体与当前综述工作的区别如表1所示。总结来看，本文的贡献如下：

- 根据 VLA 发展过程中的特点，本文将 VLA 发展历程划分成 3 个阶段：萌芽阶段，VLA 概念尚未形成，但已经出现相似功能的模型；探索阶段，VLA 模型架构“百花齐放”，但逐渐确立了以 Transformer 为核心的可扩展骨干结构；快速发展阶段，模型架构从单层往多层方向发展，并且随着数据积累，多模态 VLA 模型已经

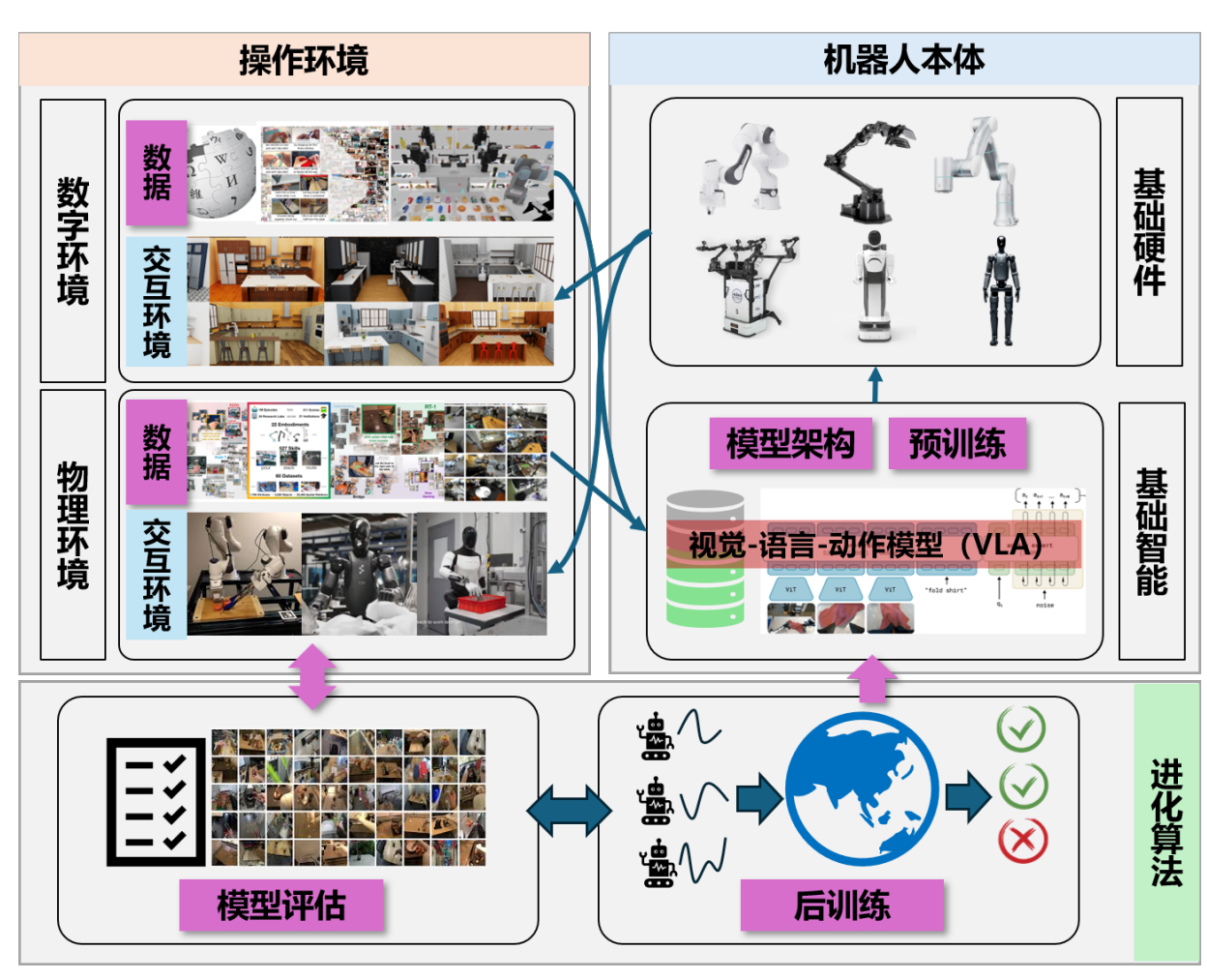

图 1: 具身操作
Fig. 1: Embodied Manipulation.

“崭露头角”。

- 根据数据类型和数据使用方式的不同，本文将 VLA 模型预训练方法划分为 4 种：单一领域数据预训练、跨域分阶段训练、跨域数据联合训练以及思维链增强。单一领域数据预训练方法是当前 VLA 常用预训练方法，但局限性比较明显，跨域数据联合训练和思维链增强具有较大的发展潜力。
- 本文将 VLA 模型后训练方法划分为 3 类：监督微调，目前 VLA 后训练的主要手段，在泛化性和持续学习能力方面面临挑战；强化微调，作为一种交互和奖励驱动的主动学习方法，具备一定的发展潜力，但尚未形成里程碑式工作；推理扩展，不需要额外的数据训练，但需要消耗时间换取性能，面临速度与性能折中。
- 本文弥补了当前综述工作中对于 VLA 模型评估工作的欠缺，从真实环境评估、仿真器评估和世界模型评估 3 个方面，全面剖析目前 VLA 模型评估现状与需求的鸿沟。

## 1　具身操作与 VLA

具身智能是一种强调智能体通过与物理环境持续交互实现感知、学习和决策的研究范式。其核心在于将智能体置于具身化（拥有物理或仿真载体）和情境化（身处动态环境）条件下，通过“感知-行动”闭环驱动智能进化。与传统 AI 处理抽象数据不同，具身智能要求智能体直接接收多模态信息输入，执行物理动作，并在环境反馈中学习任务策略。其中，环境、本体和进化算法是具身智能的 3 大核心要素。机器人操作通过控制机械臂与环境交互，完成设定的任务，相比于机器人运动和导航，该任务具有更强的环境交互和改变能力，是具身智能的典型应用场景。

传统的机器人操作通常会根据预先已知的环境和任务，对机械臂预先编程或将预训练好的模型部署到机器人上。当模型一旦部署之后，该模型的性能不会再从机器人与环境交互过程中受益，无法实现智能的迭代。因此，仍然无法脱离传统智能的范式，该范式下机器人很难处理环境和任务的变化。具身操作的核心在于当模型控制机器人本体与环境交互时，可以通过反馈的数据进一步迭代算法，从而提升模型的能力，以及对于新场景和新任务的适应能力。因此，相比于传统的机器人操作，具身操作具备更广阔的应用场景。

与具身智能相同，具身操作也包含 3 大关键部分：操作环境、机器人本体和进化算法，如图1所示。操作环境包含数字环境和物理环境。每种环境中都涵盖数据和可交互的环境。其中数据用于预训练模型构建本体的基础智能，环境用于与机器人本体交互产生新的数据。机器人本体包含基础硬件和基础智能。其中，基础硬件是实现物理交互的前提，而基础智能是机器人本体的基本能力。基础模型及其预训练构成了机器人本体的基础智能。在具身智能中，基础智能的预训练不是必须的，或者说在进化开始，机器人本体不一定具备智能，可以从零开始学习。但是在机器人操作中，从零开始学习的时间代价和物理代价是非常高的。因此，使用数据预训练获得一定能力的基础智能是合理的。在目前的具身发展中，VLA 是基础智能的核心，如何设计 VLA 模型结构，并预训练 VLA 是实现更强基础智能的关键。进化算法是具身操作中非常关键的一环，也是与传统机器人操作的核心区别。模型评估和后训练是进化算法的两大要素，其中，模型评估用于分析模型进化方向的优劣，而后训练是模型在交互过程中提升基础智能的关键。

从上述的分析中可以发现，VLA 的模型结构、训练数据、预训练、后训练和模型评估是具身操作中的核心功能模块。因此，本文内容安排如下：第 2 节对 VLA 从萌芽阶段到目前快速发展阶段的发展历程进行详细介绍；第 3 节对当前主流 VLA 模型进行解构，分析目前 VLA 模型架构特点；第 4 节介绍了 VLA 模型训练过程中使用到的数据集；第 5 节和第 6 节分别介绍了 VLA 模型的预训练与后训练方法；第 7 节对目前 VLA 模型的评测方法进行了分类和分析；第 8 节阐述了目前 VLA 模型面临的挑战和未来潜在的发展方向；最后在第 9 节中对全文进行了总结。

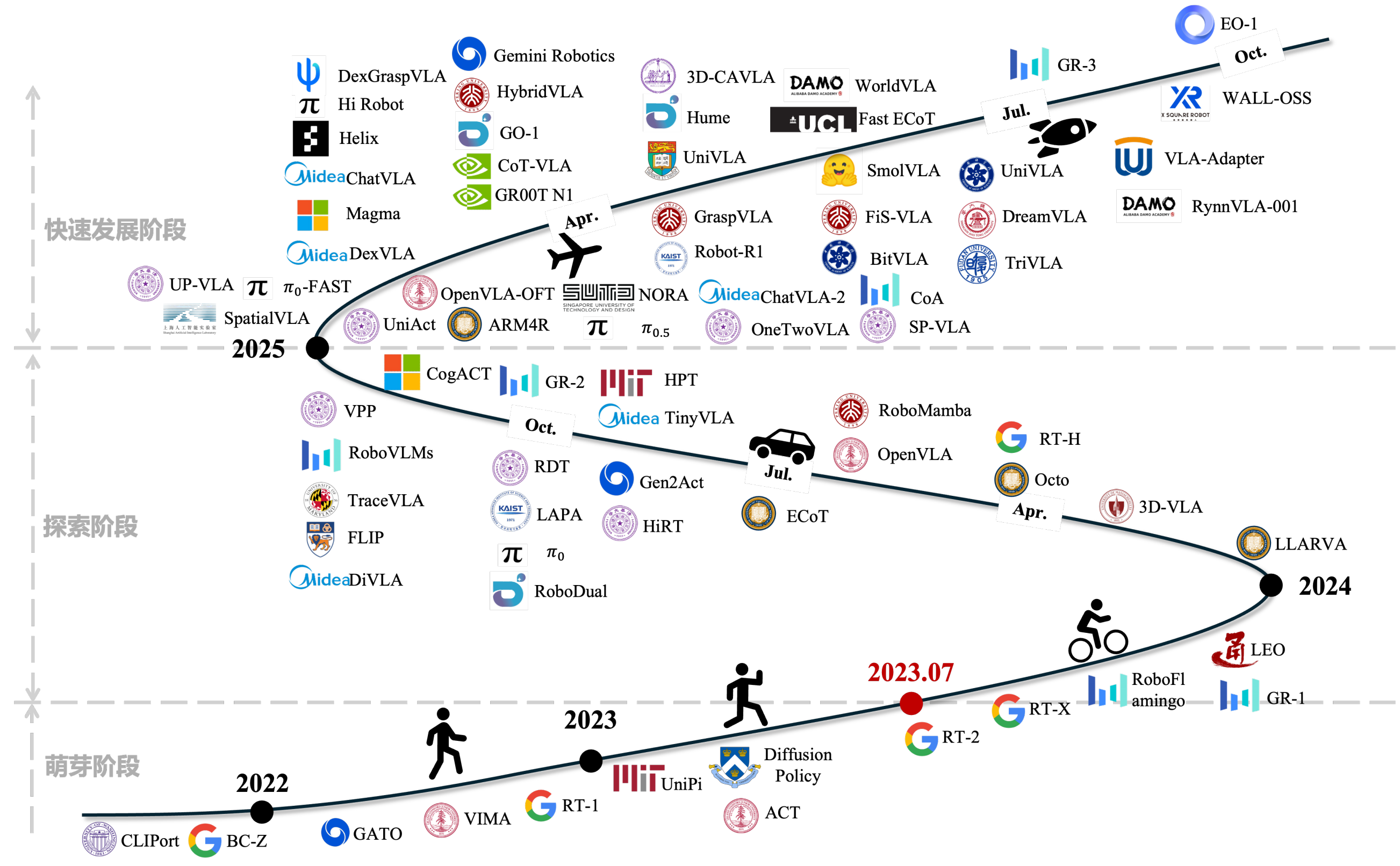

图 2: VLA 模型发展时间线

Fig. 2: The Timeline of VLA Models

## 2 VLA 的发展历程

根据视觉图像计算机器人控制量，不仅可以获得更丰富的语义信息，而且也一定程度上降低了传感器成本。经典视觉模仿学习通常是针对某个指定任务进行学习，任务之间很难实现泛化，从而限制了机器人应对开放环境和开放任务的能力。自然语言作为人类日常生活沟通的媒介，为处理开放环境与任务提供了可行的原语。早期的方法主要是使用卷积神经网络（Convolutional Neural Network, CNN）或递归神经网络，通过语言作为任务标识实现了多任务的机器人模仿学习。但是这类方法一方面由于网络结构导致的容量限制，很难在大规模任务上学习到良好的性能；另一方面，这类方法通常仅针对固定的动作原语，对于开放语义的理解与泛化比较困难。

随着 Transformer 框架的诞生与发展，大模型展现了优秀的自然语言理解和图像理解能力，并且验证了通过叠加 Transformer 模块和增加数据规模可以带来性能的提升这一扩展法则。因此，通过将大模型技术用于解决机器人控制问题，形成了目前主流的 VLA 模型：即在 Transformer 框架基础上，根据视觉图像和自然语言的任务描述作为输入，通过理解开放语义和图像内容，计算机器人控制动作。VLA 模型的概念是在 2023 年 7 月份左右首次提出[11]，目标是构建一个通用策略，由可扩展的网络结构构成，输入为视觉和语言，输出为机器人控制动作的模型。其中，语言作为一种通用的任务描述，可以适用于开放场景下的多样性任务需求。本文对“视觉”的概念进行扩展，不再局限为彩色图像，而是包含机器人系统中可能出现多种传感器信号，例如深度图像，激光雷达点云，触觉或力信号等，这种扩展定义在目前的研究工作用也被广泛使用[12-14]。从早期萌芽阶段至今，VLA 模型发展大致经历了 3 个阶段，如图2所示。

### 2.1 萌芽阶段

早期端到端视觉控制策略常采用独热编码（one-hot encoding）实现多任务区分，然而此类方法局限于封闭任务集合，难以适应开放环境中的动态需求。鉴于语言作为人类认知交互的核心媒介，将其引入任务描述可显著增强系统对复杂场景的适应性[15]。为实现策略对语言语义的理解，现有研究提出多种技术路径：通过语义解析模块提取图像中语言描述相关的物体特征，将指令与视觉模仿学习融合实现柔性抓取[16]，但该方法存在目标检测依赖性与泛

化能力受限问题；进一步改进方案 CLIPort[17] 框架采用双通路架构，结合 CLIP[18] 的语义表征与 Transporter 网络[19]的空间推理，显著提升语义理解广度与操作精度。随着预训练大模型的发展，利用预训练语言嵌入编码任务描述，经 FiLM[20] 结构融合 ResNet 视觉特征预测机械臂轨迹可以更好地处理开放语义，并在构建的百任务级操作数据集上验证了多任务学习的跨任务泛化能力[21]。

考虑到 Transformer 是用于处理序列数据的模型，如何将机器人控制问题的输入与输出序列化成为 VLA 模型的核心问题。例如，早期的代表性工作 RT-1[22] 使用 EfficientNet[23] 和语句编码器 (Sentence Encoder)[24] 分别作为图像和语言任务描述的分词器，并使用等分离散化对动作进行分词。VIMA[25] 借鉴了多模态大模型的结构，在输入任务描述中实现语言和图像的穿插，从而可以表达更复杂的任务。GATO[26] 建立了一种更统一的框架，使用一套模型和参数，可以同时完成包括机器人控制、游戏控制、文本生成、图片理解等多个任务。这些方法大部分都是从大模型的角度去考虑如何将机器人控制问题转换成序列处理问题，并没有针对机器人的特性进行设计。

在大模型中输出的令牌（Token）通常是离散化的，为了使机器人任务适应该框架，通常是把动作做离散化预处理。虽然离散化动作更符合大模型训练模式，但是一定程度上限制了机器人的执行精度。此外，每次推理动作的不一致性会导致机器人执行过程的抖动现象。针对这个问题，ACT[27] 提出了一种动作分块的思路，模型每次推理多个时刻动作，在执行过程中通过加权多个预测动作实现更平滑的运动。随着训练数据规模增加，机器人动作在相同观测下可能存在多种选择，使得训练数据集中机器人的轨迹呈多模态分布。由于确定性连续动作使用了高斯分布假设，很难建模这种行为多模态性。随着以扩散模型为代表的生成式模型在图像和视频生成领域的发展，扩散模型被用于建模机器人策略[28-30]，并取得优异的效果。这种方法在之后的工作中也被广泛的使用[31-32]。

### 2.2　探索阶段

2023 年 7 月份左右，VLA 的概念被首次提出来[11]，并且同时推出了参数量为 55B 的 VLA 模型 RT-2[11]。该模型在真实机器人上展现出优异的泛化能力，推动 VLA 进入快速发展阶段。此外，公开且统一的数据基础是 VLA 模型发展的动力之一，谷歌的研究人员整合了来自 21 个不同机构的机器人操作数据，构建了包含 1M+ 机器人轨迹的大规模数据集 OXE[33]，涵盖 22 种不同的机器人，527 种技能和 160k 个任务。该工作为机器人学习社区提供了一个协作平台，使得不同机构可以共享数据和资源，推动跨形态机器人学习这一重要研究方向的发展。在这一阶段，Transformer 成为 VLA 骨干模型的主流选择，但是对于其权重是否继承自 LLM 或 VLM 预训练权重，是一个需要讨论的话题。

早期的一些 VLA 方法秉承 GATO 的思路，通过大量的任务数据训练模型获得良好的泛化能力。相比于 RT-2[11]，Octo[31] 被设计成一种参数量更小的 VLA 模型。该模型使用轻量级 CNN 作为编码器处理图像，并且设计了块级掩码注意力机制，可在微调时添加新的观测模态而不影响预训练权重。虽然参数量比较小，但是在 WidowX、UR5、RT-1 机器人上超越 RT-1-X[33]，并且表现出与拥有 55B 参数的 RT-2-X[33] 相当的性能。更重要的是，该工作开源了预训练权重和微调脚本、完整的预训练管道和数据加载器，为 VLA 社区提供了易于使用的工具和基础设施。为了让模型更好地处理异构信息，HPT[34] 采用了包含编码器（Stem），共享主干（Trunk）和解码器（Head）的 3 层结构，可以适用于不同结构的机器人。不同于 Octo 中从头开始训练 CNN 编码器，HPT 允许使用预训练的视觉编码器，并且编码器将不同的机器人本体与视觉特征映射到统一空间中实现 Transformer 主干共享，实现了跨机器人和任务的预训练。考虑到双臂操作任务的多模态动作分布问题，RDT-1B[32] 使用改进的 Diffusion Transformer[35] 作为骨干以缓解多模态动作分布下的平均效应。该工作使用预训练的视觉编码器 SigLIP[36]，并构建了包含 1M 条轨迹的数据集，以支撑从零训练骨干网络。随着更多的机器人数据集的发布，使用大规模轨迹数据预训练机器人基础模型在机器人多任务操作上取得了显著的进步。但是相比于 LLM 和 VLM，目前机器人的数据量仍然处在非常小的量级，所训练出的机器人基础模型仍然没有达到 LLM/VLM 相当的泛化能力。因此，如何使用大规模互联网数据或继承预训练 LLM/VLM 的能力，成为进一步提升 VLA 泛化能力的关键。

虽然 Transformer 框架的可扩展性带来的模型参数和容量提升使机器人可以使用同一套策略完成不同的任务，但是模型参数的增加随之而来的是对数据量需求增加。相比于自然语言和图像数据，机器人领域的数据目前难以匹配其规模。因此，仅依靠机器人数据从头开始训练的 VLA 模型很难理解复杂的任务描述和图像中个体之间的空间关系。为了减小训练过程中对机器人数据的需求，提高 VLA 模型的理解能力和泛化能力，继承大语言模型或者视觉语言模型权重，并在此基础上使用机器人数据训练成为 VLA 模型一种常见的选择。例

如谷歌 DeepMind 推出的 RT-2[11] 继承了 PaLM-E[37] 的权重，字节跳动提出的 RoboFlamingo[38] 继承了 Flamingo[39] 的权重，斯坦福大学提出的 OpenVLA[40] 和清华大学提出的 CogACT[41] 继承了 LLaMA[42] 的权重，物理智能公司（Physical Intelligence）推出 $\pi_0$[43] 继承了 PaliGemma[44] 的权重。RoboVLMs[45] 中讨论了继承不同大模型权重对 VLA 性能的影响。

考虑到 VLA 模型训练数据不足的问题，除了继承预训练大模型的权重，也有工作开始探索如何利用视频数据来提升 VLA 模型的操作性能。最直接方法是通过结合视频生成模型的最新进展，构建面向机器人操作的视频生成模型，生成机器人操作视频序列，然后再通过机器人动作数据训练的逆动力学模型解算视频序列对应的动作[46-47]。但是这种方法对视频生成质量要求比较高。另外一种方式是通过使用大规模人类操作活动数据对模型进行预训练，再使用少量机器人轨迹数据微调获得机器人策略[48-49]。此外，通过无监督学习的方式在视频中获取潜在动作，为图像构建潜在动作标签以预训练 VLA 模型，可以将不同构型的机器人甚至是人类活动数据扩展到 VLA 训练数据范围中，然后再通过少量数据将潜在动作对齐到真实动作空间中，从而大大减少对于机器人轨迹数据的需求[50]。

### 2.3 快速发展阶段

从 2024 年底开始，VLA 模型进入快速发展阶段。大部分的模型似乎更倾向于选择预训练的 LLM/VLM 作为模型基座构建 VLA 模型。虽然这些 VLA 继承了 LLM/VLM 的权重，但是其展现的泛化能力相比于大模型仍然相差甚远。该问题的原因可能是机器人轨迹数据与文本数据和图像数据之间的错位，模型会在新的任务上过拟合而并非源于模型容量限制，从而遗忘原先具备的语言理解或多模态对齐能力，引发微调后的大模型面临“虚假遗忘”问题[51]，导致对于场景理解和泛化能力急剧下降。如何解决 VLA 模型的泛化性问题成为这一阶段的重点研究方向之一。

VLA 的泛化能力涉及很多维度，其中提高任务和操作对象的泛化能力的关键在于提升 VLA 的语义理解能力。考虑到 VLM 和 VLA 在任务场景理解和动作执行任务上的能力边界，分层架构的 VLA 模型在这一阶段成为解决复杂操作问题的热门选择。这种分层的架构将场景理解与动作执行进行隔离[52-54]，上层策略（System 2, S2）使用预训练 VLM 模型提取环境中的关键信息，实现场景理解并将子任务发送给下层策略；相比于上层策略，下层策略（System 1, S1）是一个专注于简单任务的小模型，它接收上层指令和机器人本体信息，计算当前所需控制量。除了这种常见的双层架构，也有研究工作提出了一种 3 层结构的 VLA 模型[55]，通过在 S2 和 S1 之间嵌入视频预测模型作为运动感知模块（System 3, S3），从而弥补 VLA 方法忽略动态信息的问题。虽然分层可以使 VLA 模型更好地理解环境，如何在层间传递信息是一个非常关键的问题。语言作为一种天然的信息媒介，可以用于该场景的信息传递[53,56]，并且具有良好的可解释性，但是信息的离散化也一定程度上丢失了任务与环境信息。因此，有些研究工作则采用隐空间向量的方式隐式传递信息[57-59]，同时在训练时也可以传播梯度。此外，机械臂操作轨迹作为一种机器人容易理解且具有物理意义的表达方式，也被用来作为分层 VLA 的层间信息传递[52,60-61]，从而让下层策略更容易生成跟随上层轨迹的动作。从 2025 年初，以分层架构为代表的新一代 VLA 模型逐渐发挥优势，这种架构带来的额外好处是下层策略比较灵活，可以实现比较高的推理速度，从而满足机器人控制的实时性要求[52]。

另外一种提升 VLA 泛化性的方向是从 VLA 预训练的角度出发，从模型根源上解决泛化能力问题。考虑到 VLA 所继承的 VLM 权重是通过大量的图文数据进行预训练，在训练 VLA 的过程中混入图文数据，可以激发预训练 VLM 潜能，从而提升 VLA 的视觉泛化能力。这种联合训练过程中使用的图文数据有两大类：一类是场景理解数据，例如目标检测、图像描述以及视觉问答等数据。在训练过程中一般是以少量的比例与机器人轨迹数据混合在一起训练 VLA 模型，使 VLA 保持原有图文数据理解能力的同时，提升机器人操作的泛化能力。另外一类是推理数据，通过分解任务执行过程[62-63]，或者描述有利于任务完成的场景[64]，在当前机器人轨迹数据上构建思维链，训练 VLA 模型时让模型同时输出思维链过程和动作，从而激发 VLA 模型环境理解能力。

在此期间，VLA 也向更多的维度发展。基于大规模视频数据预训练的模型也得到了进一步发展[65-66]，利用视频中的动力学信息辅助 VLA 训练可以减少对于真实机器人轨迹数据的需求[67-69]。此外，如何融入更多维度的传感器数据，让 VLA 模型更好的理解环境，是进一步提升操作能力的研究方向之一。例如利用空间信息，提升 VLA 的空间理解和操作泛化能力[70-71]，利用触觉[14]和力[13]信息提升 VLA 在精细操作任务中的能力。考虑到机器人操作的实时性要求，如何提高 VLA 推理效率是该模型在部署过程中面临的关键问题，目前的已经有研究从模型精简[72]，模型量化[73]，高效推理[74-75]等方面开展相关的研究工作。

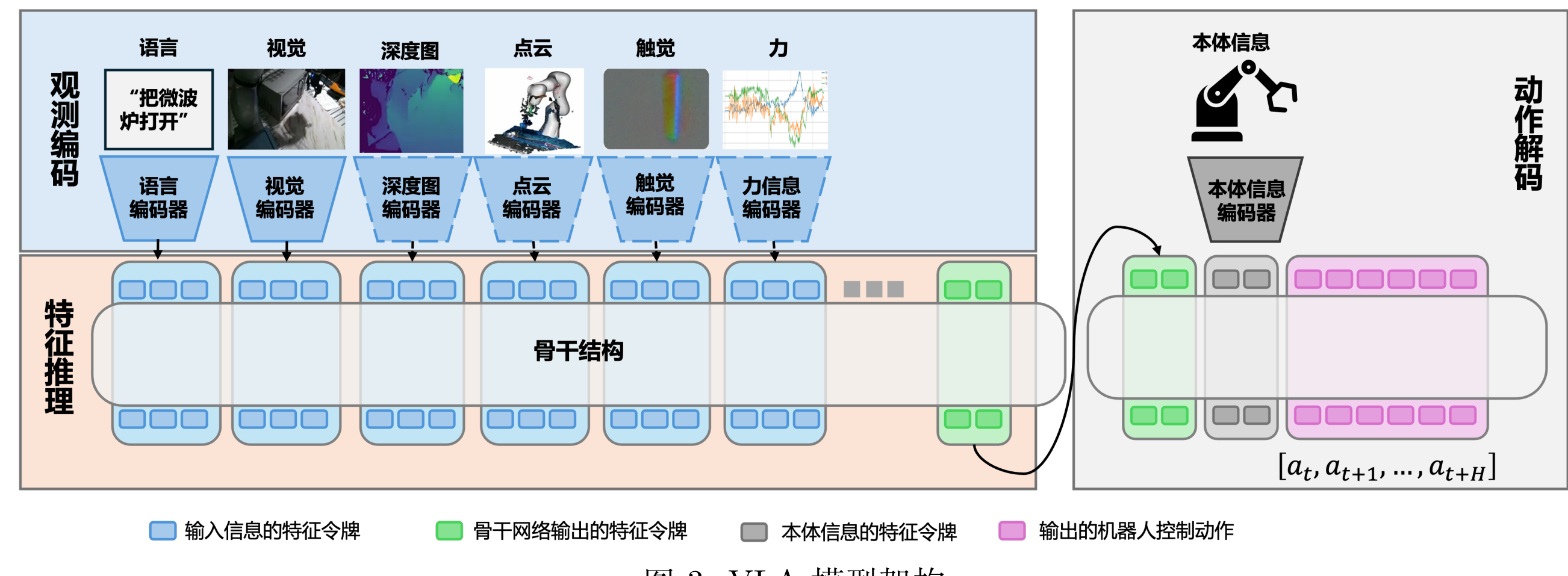

图 3: VLA 模型架构

Fig. 3: The Framework of VLA Models

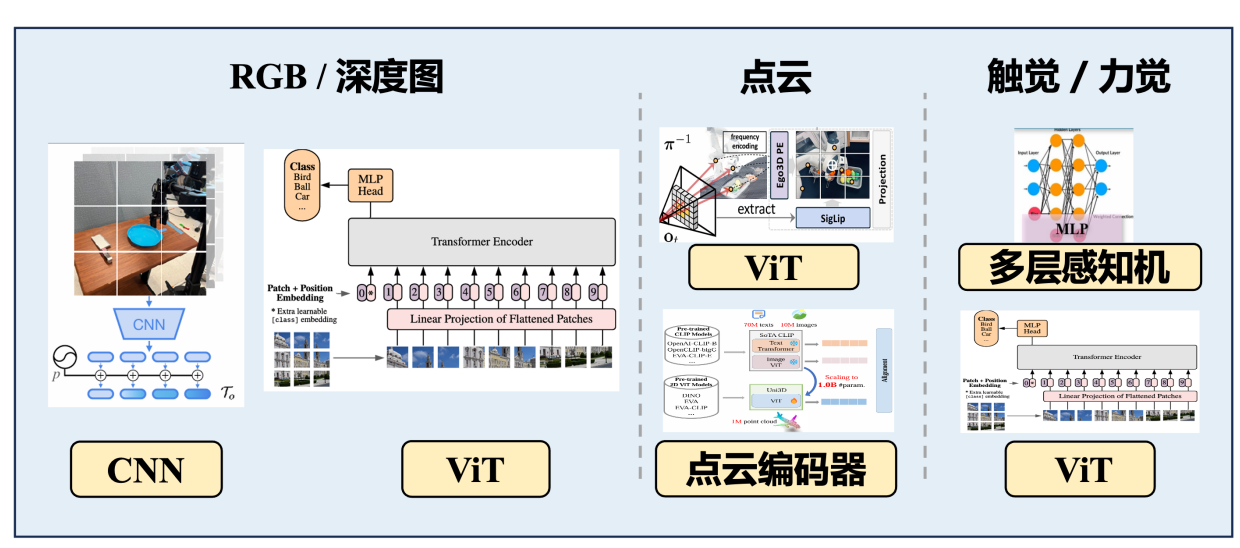

图 4: 观测编码

Fig. 4: Observation Encoder

## 3　VLA 模型架构

目前主流的 VLA 模型架构可以大致分为 3 个部分：观测编码、特征推理和动作解码。其中观测编码负责将任务描述和观测信息编码成骨干网络可以处理的观测令牌。该观测令牌经过骨干结构进行特征推理获得有助于动作决策的特征令牌，然后该特征令牌再经过动作解码获得机器人可以执行的动作。详细的介绍如图3所示。

### 3.1　观测编码

当 VLA 模型获得任务语言描述和观测信息时，观测编码的作用是将该信息转换到特征空间中。VLA 模型中涉及的观测主要是机器人中捕获的传感器信息，包括彩色图、深度图、点云、触觉和力等信息，如图4所示。对于语言描述，通常使用预训练的语言编码器 T5 或者字节对编码器（Byte Pair Encoding, BPE）作为语言的观测编码。对于观测，通常是将原始传感器观测编码到特征空间中。以图像与语言指令作为输入模态的架构长期占据主导地位，观测编码器通常被称为视觉编码器，其选择也有多种。

#### 3.1.1　针对彩色图像的观测编码

早期工作主要采用基于 CNN 的架构，如 Octo[31] 使用简单的 CNN、CLIPort[17] 使用 ResNet-50、RT-1[22] 采用 EfficientNet-B3 和 ACT[27] 采用 ResNet-18，这些模型普遍使用在互联网数据上预训练的视觉骨干网络作为图像特征提取器。这类 CNN 结构因其在传统视觉任务中的表现稳定，被迁移到机器人操作任务中。然而，随着 Vision Transformer[76]（ViT）在图像理解任务中的表现逐渐超过 CNN，VLA 模型也逐步转向基于 ViT 的视觉编码器。早期，使用 PaLI-X 视觉编码器的 RT-2[11] 模型曾在多个任务中取得良好表现。随后，结合 DINOv2[77] 和 SigLIP[36] 的联合表征逐渐成为一种趋势[40,78]。此外，也有一些工作开始探索使用 VLM 的视觉编码器进行表征，例如 $\pi_0$[43]、$\pi_{0.5}$[79] 和 $\pi_0$−FAST[80] 等使用 PaliGemma[44] 的视觉编码器在多个应用场景中取得了显著的效果。这些模型通常采用在大规模图像-文本对数据上预训练的视觉模型，并通过冻结、微调或共同训练的方式，将其接入 VLA 模型。相较于传统 CNN，ViT 模型具备更强的语义理解能力、跨模态对齐能力和更好的泛化性能，使得机器人能够更准确地解析复杂语言指令并据此执行操作。此外，使用预训练视觉编码器还能显著减少训练时间和数据需求，并提升模型零样本泛化能力。

#### 3.1.2　针对几何信息的特征编码

然而，使用单视图或多视图二维模态进行表征会导致模型缺乏空间感知能力，而这一能力在精细操作任务中非常关键[81-84]。为了增强空间感知能力，VLA 模型通常会使用多视角图像、深度图或点云获取几何信息。例如，3D-VLA[12]、FP3[85]、

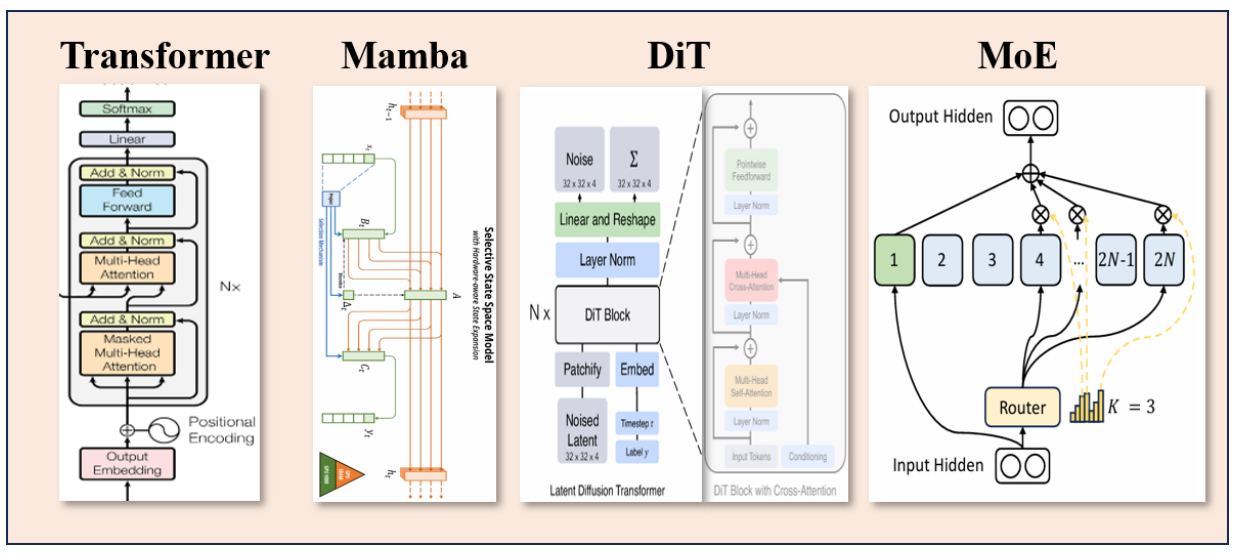

图 5: 特征推理

Fig. 5: Feature Reasoning Backbone

PointVLA[81]、CL3R[82]、QDepth-VLA[83] 等方法尝试通过直接编码三维数据来获取更多的空间信息。尽管这些模型在特定任务中取得了显著的性能提升，但由于具身三维数据在数量和质量上的限制，难以训练出一个能够与二维图像或语言编码器有效对齐的三维数据编码器，这对基于三维模态 VLA 模型的泛化能力构成了重大挑战。为了解决三维模态在数据层面的不足，一些方法如 OG-VLA[86]、BridgeVLA[87]、SpatialVLA[70]、Lift3D[88] 提出使用二维数据编码器辅助机器人理解场景中的三维空间关系，尝试将机器人获得的丰富二维语义感知提升到三维空间感知层面。其中 SpatialVLA 引入了 Ego3D 位置编码机制，将三维空间位置编码信息与利用二维数据编码器提取的二维语义特征相结合，以捕获三维场景信息。而 OG-VLA 和 BridgeVLA 则借鉴了 RVT[89]/RVT-2[90] 的思想，通过将三维点云投影到预定义的正交投影图像上，利用成熟的二维表征编码器来表征空间信息。通过将三维数据转换为二维投影图，并利用强大的二维视觉编码器进行处理，这些方法巧妙地规避了三维编码器的数据瓶颈限制，有效增强了模型的空间表征能力。

#### 3.1.3 针对交互反馈信息的特征编码

除了视觉表征外，现有研究还尝试将触觉和力/力矩这种交互反馈信号模态引入 VLA 模型，以提升机器人在复杂操作任务中的表现。尤其在精细操作或动态环境中，触觉能够为机器人提供视觉所无法捕捉的物理交互反馈，从而更准确地理解物体状态并实现精细控制。例如，TLA[91] 和 VTLA[14] 通过在触觉模态上微调 Qwen-VL 的 ViT 模块，从视触觉图像数据中提取有效的特征。由于视触觉传感器可以输出图像形式的触觉数据，机器人能够借助预训练视觉编码器强大的表示能力，实现更高效的跨模态迁移，从而提升在如物体区分、滑动检测等精细操作任务中的性能。相比之下，其他方法则直接处理原始的力或触觉信号。例如，Tactile-VLA[92] 针对其高分辨率触觉传感器采集的法向力和剪切力数据，采用了一个多层感知机作为编码器。该编码器将一段时间内的连续触觉测量值处理成一个融合的特征令牌，然后在输入层面与视觉和语言令牌进行拼接，实现多模态信息的深度融合。将触觉对齐到 VLA 系统中需要依赖丰富且大量的触觉数据，文献[93]详细介绍了近年来触觉数据生成方法。而 ForceVLA[13] 针对六维力/力矩信号，通过一个简单的线性投影将其编码为力嵌入特征，并引入混合专家（Mixture of Experts，MoE）结构，实现视觉、语言与力信号的深度融合。在需要复杂物理交互的任务中，这种显式引入力/力矩信号的方式显著提升了模型的精细操作能力。然而，类似于三维模态，触觉与力觉模态同样面临获取大规模、高质量和多样化数据集的挑战。与当前的相对成熟的图像与语言数据形式相比，触觉/力信号的表示方式尚未统一。尽管如 AnyTouch[94] 提出了一种统一四类视触觉传感器数据的对齐框架，但在更大规模场景下进行训练和泛化仍存在较大障碍。因此，如何实现触觉/力模态在感知模型中的标准化、数据共享与可扩展性，仍是具身多模态大模型研究中的关键难题之一。

### 3.2 特征推理

在获得不同时刻的观测编码和任务描述编码后，需要设计骨干网络捕获特征之间的相关信息，并推理出适用于预测机器人动作的特征。对于特征推理的骨干网络结构，目前也有多种选择，如图5所示。作为一种基于自注意力机制的深度学习架构，Transformer 被广泛应用于自然语言处理和其他序列数据处理任务中，展现出强大的可扩展能力。RT-2 首次将其引入到 VLA 模型中，通过自注意力机制捕捉输入序列中各元素之间的依赖关系，使得视觉、语言和动作信息能够在同一架构中得到高效处理和深度建模。此外，Transformer 的层次结构和位置编码使其能够在多模态任务中捕捉长程依赖关系，从而提升模型在复杂任务中的表现。

#### 3.2.1 基于 Transformer/DiT 的特征推理

Transformer 在建模语言指令和观测到动作的映射时，通常建模为确定性映射。当训练数据集中存在多模态数据分布时，这种确定性映射会学习到多个动作模式的”平均值”，尤其是在动作多模态性更强的双臂操作任务上，即同一任务或指令在不同场景或条件下可以对应多种不同的动作执行方式，这种平均可能产生完全不可行的动作。扩散模型作为一种基于逐步添加噪声和去噪过程的生成模型，通过一系列噪声添加和去噪步骤，能够更好地拟合复杂的数据分布，捕捉动作分布的完整多模态性。Diffusion Transformer[35]（DiT）将扩散模型与 Transformer 结合，利用强大的自注意机制建模长

程依赖，能够更好地处理全局和局部信息，从而生成更高质量的样本。该模型也被用于作为 VLA 骨干网络。RDT[32] 提出使用 DiT 代替常规的 Transformer，使用 QKNorm 和 RMSNorm 解决大规模数据训练时的数值不稳定和梯度爆炸问题，并使用多层感知机代替线性解码器，更好地近似非线性机器人动作。此外，该工作提出了交替条件注入，可以处理高维变长的图像和语言条件。GR00T N1[58] 和 TriVLA[55] 都使用 DiT 作为 S1 的主干结构，将来自 VLM 的高级、抽象的输出，高效地转化为机器人可以执行的、流畅且精确的物理动作。

#### 3.2.2　基于 MoE 的特征推理

在 VLA 模型中，骨干网络为了对齐视觉，语言和动作 3 种模态，需要经过大量且多样的文本、视觉和控制等任务数据训练。在常规的 Transformer 结构中，所有的样本和任务都共享同一组参数，不仅限制了模型的表达能力，而且任务之间容易出现干扰或者任务遗忘现象。MoE 作为一种由多个独立神经网络“专家”构成的可扩展的结构，通过动态路由选择机制为不同的任务或样本分配参数共享空间，可以实现更大的模型容量和更强的泛化能力。在 VLA 训练过程中，由于继承的 VLM 权重大部分是由图文数据进行预训练的，直接使用机器人轨迹数据继续训练会损害原始预训练权重，产生“虚假遗忘”现象，从而导致模型的泛化能力下降。使用图文数据和机器人轨迹数据联合训练可以一定程度上缓解上述问题，但是在 Transformer 结构中，机器人控制任务和图文理解任务在共享参数空间中会相互竞争，从而导致联合训练时两种任务的性能都下降。ChatVLA[51] 提出在 Transformer 中使用 MoE 替换原始的前馈网络模块（FeedForward Network, FFN），减小任务之间的干扰，并通过共享注意力机制，实现跨任务知识迁移。但是这种静态专家选择机制仍然会遗忘之前训练获得的知识。ChatVLA-2[95] 在前述工作的基础上，引入了动态混合专家架构，通过自适应地选择专家模块，保持特征贡献，避免破坏原始的 VLM 结构，从而缓解了知识遗忘问题，显著提升开放世界推理和泛化能力。

#### 3.2.3　基于 Mamba 的特征推理

虽然基于 Transformer 的 VLA 展现良好的性能，但是当输入令牌数量增加时，其推理所需计算量随序列长度平方增长，并且推理时需要缓存的历史参数，内存消耗与序列长度呈正比，大大限制了其在机器人这种推理实时性要求高，有限计算资源场景下的应用。Mamba[96] 作为一种基于状态空间模型 (State Space Models, SSMs) 的神经网络架构，通过引入输入依赖的选择性机制，避免无关上下文干扰，

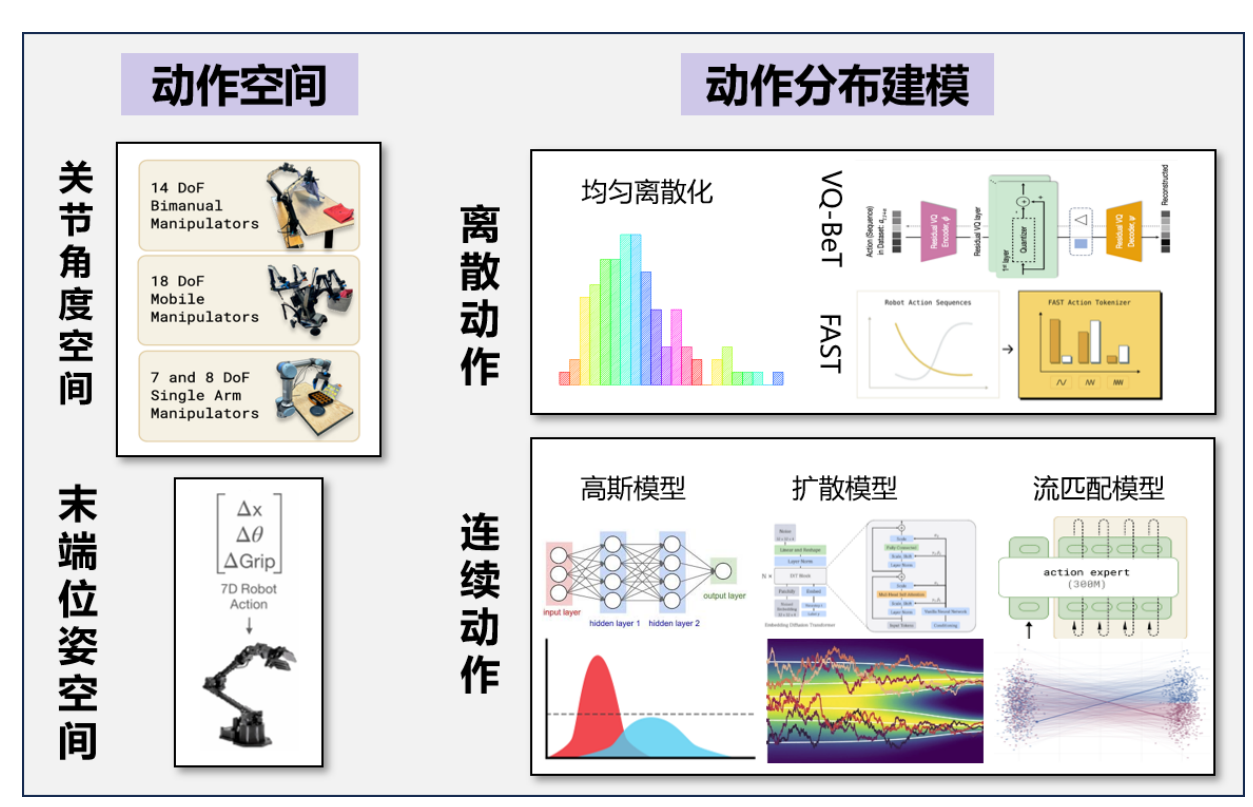

图 6：动作解码

Fig. 6: Action Decoder

实现了推理时的线性计算复杂度，从而获得比同规模 Transformer 更快的推理速度。RoboMamba[97] 将 Mamba 作为 VLA 的骨干，相比于 Transformer，Mamba 具有线性推理复杂度，大大提高了 VLA 的推理速度。并且具有更强的上下文感知推理能力，适合处理机器人的多模态序列数据。

### 3.3　动作解码

输入令牌在经过骨干模型推理后输出特征令牌，然后通过解码器转换为指定模态。在 LLM 和 VLM 中，大模型解码出来的模态是自然语言。而对于 VLA 模型，需要解码出来机器人可执行的动作。如何选择动作空间以及如何建模动作分布则是动作解码面临的核心问题。

#### 3.3.1　动作空间选择

在目前的 VLA 模型研究工作中，常用的动作空间有两类，如图6左半部分所示。一类是机械臂末端姿态的相对变化量[11,31,40]，这类动作空间通常依赖当前机械臂的位置和姿态做为基准，推理下一时刻机械臂末端需要到达的位置和姿态，再通过机械臂的逆动力学模型解算出关节所需要的控制量，最后通过关节角度控制器驱动机械臂运动。这种动作空间的优势在于可以将不同形态机械臂动作空间统一，从而实现跨形态机械臂数据的混合训练。另外一类是机械臂关节角度空间[32,43]，这类动作空间通常直接输出机械臂每个关节的目标角度，通过关节角度控制器直接驱动机械臂运动，避免了对于机械臂逆运动学的需求。但是由于不同机械臂关节数量不同，机械臂长度，质量以及所使用电机性能不同，导致该动作空间下很难直接实现跨域数据利用和泛化。

#### 3.3.2　动作分布建模选择

在目前 VLA 研究工作中，对于动作分布建模方式可以分为两大类：离散动作和连续动作，如图6右

半部分所示。其中，离散动作是将连续动作按照一定的规则进行离散化。一种简单的方式是将连续动作空间的每个维度均匀划分成 256 区间[11,40]，让模型预测属于每个区间的概率。但是由于实际动作分布并不是均匀分布，会导致这种在整个动作空间内均匀离散化的方式无法在高频动作区域提供足够的精细度。SpatialVLA[70] 提出了自适应动作网格，首先通过统计整个数据集中的动作拟合高斯分布，并基于概率密度划分网格区间，专注于高频动作区域，也可以实现跨机器人的空间动作对齐，相比于传统的均匀离散化方法需要更少的令牌数量。BeT[98] 使用 $k$ 均值（$k$-means）方法对数据集中的动作进行聚类，并使用聚类中心作为离散化后的区间中心，此外为每个区间学习连续的动作偏移量。但是 $k$ 值的选择是一个关键问题，如果 $k$ 值太小，多个不同的动作模态会被分配到同一个聚类中心，导致模态信息丢失；如果 $k$ 值太大，单个模态可能被分割成多个聚类，破坏了模态的连续性。VQ-BeT[99] 提出使用残差向量量化（Residual VQ）替代传统的 $k$ 均值聚类来对连续动作进行标记化，解决了 BeT 中 $k$ 均值在高维动作空间和长序列上的扩展性问题，以及缺乏梯度信息的局限性。现有简单的均匀离散化方法在高频机器人控制任务中表现很差，高频数据中动作令牌之间的高相关性会导致边际信息量接近零，因此，FAST[80] 提出了将时序信号压缩的经典方法与现代自然语言处理中的分词方法相结合，将信息集中到少数低频系数，提高每个令牌的信息含量。实验表明该方法在 15 种不同机器人形态上都能有效工作，并且支持关节空间、末端执行器空间、不同维度的动作。

虽然离散动作与 LLM/VLM 的预测模式更为贴近，但是动作离散化会损害操作精度，并且通常需要额外的动作解码器，对于高频精细操作任务并不合适。早期的一些工作将模型提取的特征直接通过线性层或者多层感知机映射到连续动作空间，这种确定性映射模式下会学习到多个运动模式的平均值。当数据中的行为多模态性比较复杂时，这种平均可能会产生完全不可行的动作。ACT[27] 采用条件变分自编码器建模人类演示数据的随机性和多模态行为，缓解这种均匀化现象。扩散策略（Diffusion Policy）[28] 利用生成模型领域的先进技术，首次将扩散模型引入到策略建模中，利用扩散模型强大的分布建模能力，可以有效地覆盖数据中的多模态行为分布。使用扩散模型建模预测动作分布逐渐被越来越多的工作采纳[31]。$\pi_0$[43] 和 $\pi_{0.5}$[79] 使用流匹配（Flow Matching）模型，直接预测去噪向量场，避免了扩散模型复杂的噪声构建，从而实现更少的采样步数和更高的控制频率。

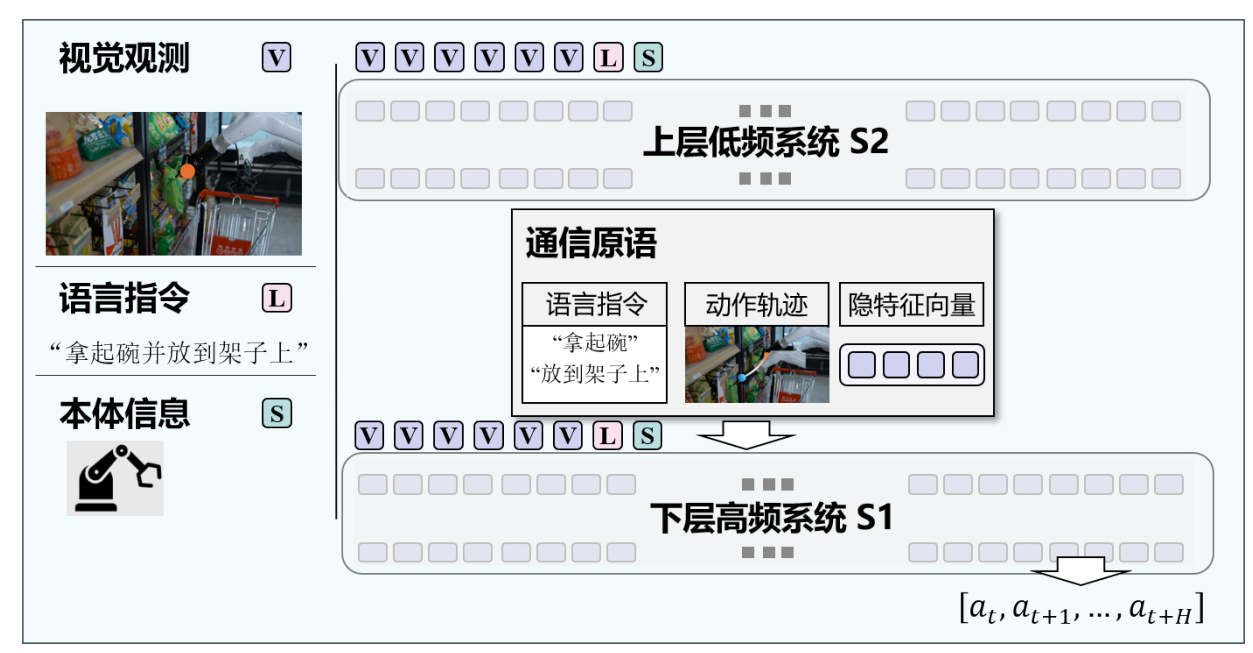

图 7: 分层系统

Fig. 7: Hierarchical System

采用离散动作空间的优势在于训练速度快，但是动作解码复杂，控制精度低；采用连续动作可以实现高频高精度的控制任务，但是训练速度比较慢。HybridVLA[100] 提出了一种融合扩散连续动作和自回归预测离散动作的方法，将扩散动作的连续性质与自回归生成的推理能力相结合，根据场景和任务自适应地选择合理的动作。物理智能的研究人员设计了一种联合训练机制[101]，利用 FAST 离散动作用于表征学习，模型在训练过程中快速收敛，连续动作用于推理时的精确控制，并且使用梯度隔离技术，阻止在连续动作训练时，从动作专家到 VLM 骨干网络的梯度流动，可以获得更好的泛化能力。

### 3.4 分层系统

前面介绍的观测编码、特征推理和动作解码构成完整的 VLA 模型，这种模型通常是由单个可扩展网络作为骨干，虽然可以处理多个时间步的观测，但是这种单层结构下的 VLA 模型很难处理长时域任务。因此，分层 VLA 架构开始出现，不同于单层 VLA 模型使用单个 Transformer 结构同时处理复杂的任务描述、多个时刻的任务观测以及多个时刻的动作输出，分层 VLA 模型将 VLA 任务拆解成长时域复杂任务理解与短时域动作生成，并使用 2 个模型分别完成相应的任务，如图7所示。对于长时域复杂任务理解，通常会使用预训练的 VLM 作为其任务理解模型（S2 系统），发挥其优秀的复杂文本理解和场景泛化能力；对于短时域动作生成，通常会借鉴之前的单层 VLA 模型（S1 系统），发挥其灵活的动作生成能力。在单层 VLA 模型中也有使用单独的模型生成动作[43]，分层 VLA 系统与其不同之处在于，短时序动作生成模型会接受图像输入，是一个相对独立的闭环系统。而单层 VLA 模型中的单独动作生成模块依赖于 VLA 输出的特征，无法独立 VLA 骨干单独运行。

不同层之间如何高效地通信是分层系统的核心，如何选择通信原语则是高效通信的关键。目前的工

作使用的通信原语大致可以分为 3 类：文本语言，动作轨迹和隐特征向量。文本语言是预训练 VLM 容易理解也擅长生成的原语类型，因此，在很多分层 VLA 系统中使用，例如 Hi Robot[53] 中有 2 个模型，其中一个负责将复杂的指令和用户反馈处理成简单的原子指令，另外一个模型将该原子指令生成执行动作。这种结构可以使的 VLA 模型处理复杂的长时序任务，并且也能够及时诊断模型理解是否出现问题。DexGraspVLA[56] 利用预训练的 VLM 通过语言模式直接生成待抓取的目标位置，大大缩减了动作生成系统的指令跟随难度，提升了 VLA 模型的语义泛化能力。虽然文本语言是一种非常方便，且容易解释的通信原语，但是对于动作生成模型，尤其是机器人动作生成模型，未来的动作轨迹则是一种更容易让模型理解方式[52,61]。这类方法一般是通过系统 S2 生成长时域动作轨迹并作为系统 S1 的条件，指导其生成更精细的动作。另外，还有一类原语是采用隐空间向量，不同于前述 2 种方式通过显式的表达将不同层的模型隔离，这种方式则是通过梯度传播的原语方式将 2 个系统连接，从而可以实现端到端训练。例如 HiRT[54] 将系统 S2 通过处理视觉和语言信息后获得的多个特征进行池化，并将其作为系统 S1 的条件。GR00T N1[58] 使用交叉注意力机制将来自系统 S2 的特征与系统 S1 更精细地融合。GO-1[65] 设计了隐空间规划作为分层系统之间的通信原语，不仅可以使模型预测长时域动作，并且可以充分利用不同领域的数据。除了典型的双层系统外，TriVLA[55] 提出了一种 3 层系统，用于解决双层系统对于环境中动态部分缺乏理解的问题。

### 3.5 挑战与发展趋势

由于数据的限制，单层 VLA 模型同时处理复杂任务理解和动作生成的能力有限，分层 VLA 模型可能会成为越来越多的选择。需要注意的是，目前的 VLA 核心架构受 LLM/VLM 发展路径启发，其所展现的局限性可能是由于机器人控制任务与文本生成任务本身的差异性导致。因此，如何将更多的机器人信息模态合理地融入到 VLA 模型架构中是一个非常值得探索的方向，从而使 VLA 模型更全面的理解世界。此外，模型结构设计会影响 VLA 模型的泛化能力和推理速度，当模型参数量比较大时，展现出的泛化性通常会比较好，但是会导致动作的输出频率比较低，从而限制其处理动态环境的能力。如何更好地处理泛化性与速度的折中，是 VLA 模型应用到真实机器人上需要解决的问题之一。此外，由于机器人上算力有限，随着 VLA 模型参数的增强，本地部署 VLA 模型的难度会越来越大，如何充分地利用云端算力，实现高效地“端-云”协同部署，也是一个值得探索的方向。

## 4 VLA 训练数据

根据大语言模型和多模态大模型的训练经验，获得面向开放环境的通用 VLA 模型需要使用海量的训练数据。相比于大模型预训练所使用的数据集，目前积累的机器人领域的数据量级相差甚远。此外，目前已有工作表明[40,43]，仅依赖机器人轨迹数据训练的 VLA 模型面临严重的泛化问题。如何扩展 VLA 模型训练数据，成为当前 VLA 领域的重点研究方向之一。英伟达的研究人员提出了 VLA 训练数据金字塔的概念[58]，将 VLA 预训练过程中使用的数据划分成 3 种类型：互联网数据，仿真数据和真实机器人采集数据。为了更好的梳理数据与训练方法，考虑到目前基于视频预训练 VLA 模型方法的发展，本文把互联网数据中的图文数据和视频数据分开，将目前 VLA 预训练数据集划分成 4 类：互联网图文数据，视频数据，仿真数据和真实机器人采集数据。具体如图8左半部分和表2所示。

### 4.1 互联网图文数据

VLA 模型控制机械臂实现操作的前提是理解环境和任务，海量的互联网图文数据能够帮助机器人构建通用知识和通用场景理解能力，为实现视觉泛化奠定基础。常用的数据集包括视觉理解任务、视觉问答任务和视觉推理任务。例如 COCO 数据集[102] 涵盖了目标检测、实例分割、关键点检测以及图像字幕生成等任务。CapsFusion[103] 是一个更大规模的高质量图像-文本对数据集，常用于多模态大模型预训练。VQAv2[104] 和 TextVQA[105] 则是面向视觉问答的大规模数据集，GQA[106] 也是一个大规模视觉问答数据，但更侧重于组合性推理和多步推理。目前很多 VLA 模型通过继承预训练的 VLM 模型权重，并没有显式地用到这些数据，但是所继承的 VLM 在预训练过程中通常会使用这些数据。

### 4.2 视频数据

在海量图文数据训练下，大模型能够理解静态环境，但是对于动态环境以及如何完成任务仍然没有先验。人类活动大规模视频数据则用于构建 VLA 模型对于动态环境理解和动作操作模式的理解基础。Something-Something V2[107] 是一个大规模视频数据集，涵盖了人类日常生活中的 174 种预定义动作。EPIC-KITCHENS-100[108] 和 Ego-4D[109] 则是以第一人称视角拍摄的大规模人类活动视频数据集，Ego-EXO-4D[110] 是在 Ego-4D 基础上增加了多视角同步捕捉，旨在推动视频学习、多模态感知和技能评估。这些视频数据集包含了丰富多样的运动技能，以及任务的完成过程，但是没包含动作信息，很

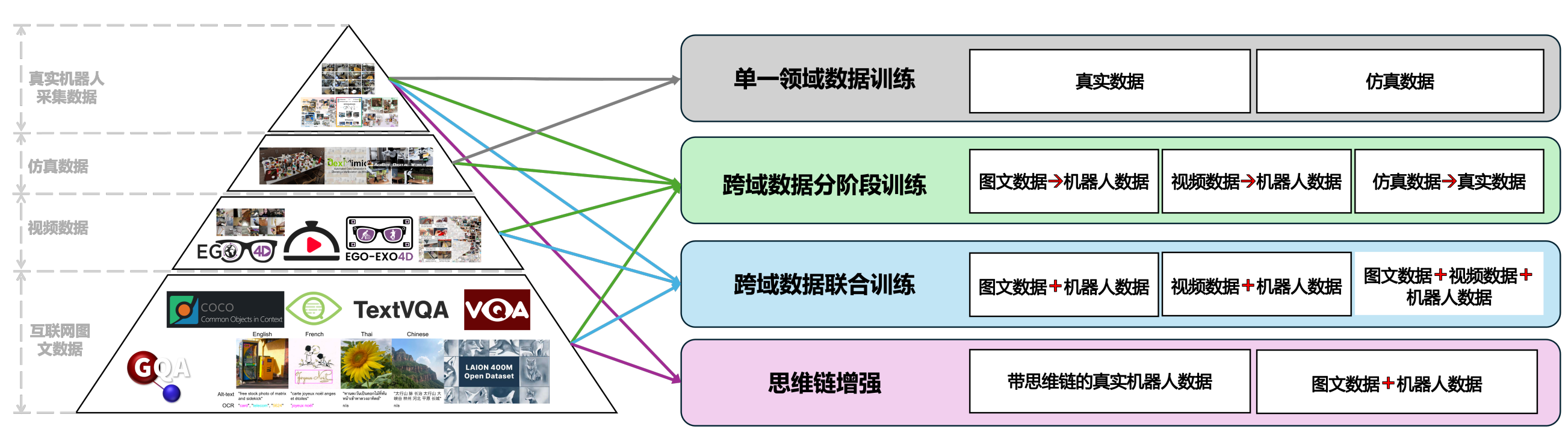

图 8: 数据金字塔和 VLA 预训练方法。图中彩色箭头表示不同的训练方法所使用的数据类型。在每个训练方法中，红色箭头表示使用数据的顺序，红色加号表示不同的数据一起使用实现联合训练。

Fig. 8: Data Pyramid and VLA Pre-training Methods. The colored arrows in the figure represent the data types used by different training methods. In each training method, the red → indicate the order in which the data is used, and the red + indicates that different data are used together to achieve joint training.

难直接用于训练动作生成模型。这类数据集通常有 3 种方式参与到 VLA 模型训练过程中：可以通过将视频预测作为 VLA 模型的辅助任务增强其对环境的感知理解能力[68,48-49]；或者使用关键点检测小模型，对视频中的手部关键点进行标注，然后映射到机械臂空间中，从而生成动作标签直接用于 VLA 模型训练[111-112]；此外，也有研究工作通过自监督学习方法学习视频中的隐含动作标注，将其作为动作标签用于 VLA 模型训练[50,65-66,113]。

### 4.3 仿真数据

视频数据中包含的时序信息有助于模型理解动态环境和交互过程，但是由于缺乏真实的机器人动作数据，模型仍然不具备理解和生成真实机器人动作的能力。考虑到真实场景的搭建成本和数据采集的人力成本，使用仿真器生成大量环境，构建带有动作标签的机器人合成轨迹数据成为了一种高性价比选择。为了保障模型的泛化能力，如何在仿真器里构建丰富的场景和多样的环境成为合成数据的关键。RoboCasa[114] 是在 RoboSuite 仿真平台的基础上构建出的一个大规模仿真框架，包含了超过 120 个逼真的厨房场景和 2500 多个高质量的 3D 物体，并提供了一个用于模型训练的大规模多任务数据集，其中包含了超过 100k 条机器生成的运动轨迹。SynGrasp-1B[115] 是针对抓取任务，在 Issac Sim 仿真平台上构建的 10B 帧规模的机器人数据集，其中包含 10k 个物体实例，涵盖 240 个类别，并采用光线追踪渲染和大范围域随机化，确保视觉和几何变化的广泛覆盖。RoboTwin2.0[116] 利用多模态大型语言模型和仿真在环的反馈机制，构建了一个自动化的数据生成流程，能够自动生成高质量、专家级别的机器人操作轨迹。该工作也发布了面向双臂操作的大规模资源，包括包含 731 个物体实例（横跨 147 个类别）的大规模物体资产库，以及包含超过 100k 条专家操作轨迹的数据集，涵盖了 50 种任务和 5 种不同的机器人形态。DexMimicGen[117] 提出了一种从极少数的人类演示中，自动为具有灵巧双手的机器人合成大量的操作轨迹的方法，并推出了一套包含 9 个不同任务的模拟环境。FetchBot[118] 针对杂乱场景下的无碰撞抓取任务，在 Issac Gym 环境中构建了一个高效的基于体素的场景生成方法 UniVoxGen，可以生成上百万种场景，并构建了一个包含数十万条轨迹的机器人操作数据。仿真数据的优势在于可以以低成本的方式生成海量的数据，但是它也面临很多的局限性，其中包括对于柔性或可变形物体的模拟保真度、复杂任务的专家轨迹生成、任务种类和场景的多样性以及仿真环境与真实环境的差异性等。

### 4.4 真实机器人采集数据

大量的仿真机器人轨迹数据可以构建 VLA 模型中场景理解与动作生成之间的关联关系，但是由于仿真环境中机器人与真实机器人动力学差异以及视觉环境差异，导致仿真数据训练的 VLA 模型很难直接用到真实的机器人和任务中。因此，构建大规模的真实场景和任务多样性丰富的机器人数据集，是实现 VLA 在真实世界和机器人上泛化的前提。加州大学伯克利分校联合斯坦福大学等机构发布的 Bridge 数据集[119]，旨在验证跨领域数据复用能否提升机器人泛化能力，为通用 VLA 模型奠定数据基础，之后对该数据进行了升级，增加了语言指令，轨迹数量和任务种类[120]。此外，斯坦福大学、加州大学伯克利分校等 13 家国际知名机构联合创建的大规模机器人操作数据集 DROID[121]，包含 76k 条

表 2: 数据集与相关方法汇总

Tab. 2: Summary of Datasets and Related Methods.

| 分类 | 名称 | 描述 | 规模 | 支持任务 | 相关方法 |
| --- | --- | --- | --- | --- | --- |
| 互联网图文数据 | CapsFusion [103] | 大规模图像-文本对数据集，为多模态预训练设计，旨在解决现有图像-文本数据集的噪声问题和低质量标注问题 | 120M 图像-文本对 | 图像描述生成，多模态预训练 | $\pi_0$ [43] |
| | COCO [102] | 大规模图像数据集，包含 80 个物体类别和 91 种材料类别，每张图片 5 个语句描述，且有 250k 个带关键点标注的行人 | 330k 图片 | 目标检测，实例分割、关键点检测，图像描述生成 | $\pi_0$ [43]，ChatVLA [51]，ChatVLA-2 [95] |
| | GQA [106] | 大规模视觉问答数据集，专注于真实世界的视觉推理和组合性问答 | 22.6M 问题，113K 图像 | 组合性推理，视觉问答 | ChatVLA [51]，ChatVLA-2 [95] |
| | LAION-400M [122] | 大规模图像-文本对数据集，包含图像 URL、图像和图像描述的嵌入、图像与描述之间的相似性评分，以及元数据 | 400M 图像-文本对 | 图文检索，图文生成，多模态预训练 | UniPi [46] |
| | PixMo [123] | 大规模图像-文本对数据集，图像涵盖 70 多个主题，每张图像描述由 3 位标注者通过语音生成 | 712k 图像，1.3M 描述 | 图像描述生成，多模态预训练 | $\pi_0$ [43] |
| | TextVQA [105] | 大规模视觉问答数据集，要求模型理解图像中的文本内容来回答问题 | 28k 图片，45.3k 问题，453k 回答 | 文本推理，视觉问答 | ChatVLA [51]，ChatVLA-2 [95] |
| | VQAv2 [104] | 大规模开放式问答数据集，由人工标注，面向开放世界视觉问答任务 | 265k 图片，443k 问题，4.43M 回答 | 常识推理，视觉问答 | $\pi_0$ [43] |
| | WebLI [124] | 超大规模多语言图像-文本对数据集，涵盖 36 种语言和多样化文化背景，包含 13 亿图像-文本对，旨在提升视觉语言模型在全球范围内的泛化能力与文化适应性 | 10B 图像-文本对 | 光学字符识别，图文检索，图像描述生成，视觉问答，多模态预训练 | RT-2 [11] |
| 视频数据 | Ego-4D [109] | 大规模第一人称视角视频数据集，涵盖数百种场景，由来自全球 74 个地点和 9 个不同国家的 931 名参与者拍摄 | 3670 小时视频 | 视频理解，多模态感知 | GO-1 [65]，GR-1 [48]，GR00T N1 [58]，Magma [125]，UniVLA [66] |
| | Ego-Exo-4D [110] | 由 Ego-4D 数据集扩展的大规模第一/第三人称视角的多模态视频数据集，增加多视角同步捕捉，专注于技能活动研究 | 1286 小时视频 | 跨视角表征学习，技能理解，多模态感知 | GR00T N1 [58] |
| | EPIC-KITCHENS-100 [108] | 大规模第一人称视角视频数据集，包含 45 个厨房环境下的动作识别，捕捉了多种家庭活动，包括 90k 个动作 | 100 小时视频，20M 帧 | 动作识别，环境理解，多模态推理，场景泛化 | ARM4R [71]，CoT-VLA [67]，GR-2 [49]，GR00T N1 [58]，Magma [125]，HPT [34] |
| | Howto100M | 大规模叙述视频数据集，主要是教学视频，其中内容创建者教授复杂的任务，并明确解释屏幕上的视觉内容 | 136M 视频片段 | 图像描述生成，多模态预训练 | GR-2 [49] |
| | Kinetics-700 [126] | 大规模视频数据集，涵盖 700 种人类动作类别，包含人与物体及人与人之间的互动 | 650k 个视频 | 动作识别，视频理解 | GR-2 [49] |
| | Something-Something V2 [107] | 大规模带标记视频数据集，包含人类使用日常物品执行的 174 种基本动作 | 220k 视频片段 | 动作识别，自监督学习，多模态推理 | CoT-VLA [67]，GR-2 [49]，LAPA [50]，Magma [125]，TriVLA [55]，VPP [47]， |
| 仿真数据 | DexMimicGen [117] | 大规模仿真数据集，涵盖涉及精密操作和灵巧手场景下多种复杂操作任务，通过人类演示与仿真生成 | 21k 轨迹 | 灵巧操作学习，精细控制，仿真到现实迁移 | GR00T N1 [58]，GR00T N1.5 [58] |
| | RoboCasa [114] | 大规模仿真数据集，提供 120 种厨房场景与 2500 个 3D 物体，结合大语言模型生成任务与自动轨迹生成，支持通用机器人操作与策略学习 | 超过 100k 轨迹 | 策略学习，环境理解，多模态预训练，仿真到现实迁移 | GR00T N1 [58] |
| | SynGrasp-1B [115] | 大规模合成动作数据集，专注于机器人抓取技能的学，涵盖 240 个物体类别和 10k 个物体 | 1B 帧 | 抓取策略学习，仿真到现实迁移，跨任务泛化 | GraspVLA [115] |

| 分类 | 名称 | 描述 | 规模 | 支持任务 | 相关方法 |
|---|---|---|---|---|---|
| 真实机器人数据 | AgiBot World[65] | 大规模多场景数据集，涵盖家居、餐饮、工业、商超及办公 5 大核心场景，涵盖超过 100 种真实场景和 3000 多种日常物品，其中 80% 的任务为长程任务 | 1M 轨迹，2976.4 小时交互数据 | 多任务学习，跨场景泛化，多模态预训练，仿真与真实结合训练 | GO-1[65]，GR00T N1[58] |
| | BC-Z[21] | 大规模机器人模仿学习数据集，涵盖 100 种操作任务，通过专家远程操作与自主收集，支持零样本任务泛化和语言与视频条件下的策略学习 | 25.8k 轨迹 | 多任务学习，跨场景泛化，多模态预训练 | $\pi_0$[43]，CoT-VLA[67]，TraceVLA[127]，UniPi[46] |
| | Bridge Data[119] | 大规模多任务操作数据集，涵盖使用 WidowX 机械臂在 10 个环境中收集的 71 个厨房任务 | 7.2k 轨迹 | 多任务学习，跨场景泛化，多模态预训练 | UniPi[46]，GR-2[49] |
| | Bridge Data V2[120] | 大规模多任务操作操数据集，使用 WidowX 机械臂在 24 个环境中收集，涵盖广泛任务与环境变化，支持图像和语言条件下的多任务学习与技能泛化 | 60k 轨迹 | 多任务学习，跨场景泛化，多模态预训练 | $\pi_0$[43]，ECoT[128]，LAPA[50]，NORA[129]，RDT[32]，TraceVLA[127] |
| | DROID[121] | 大规模真实机器人操作数据集，覆盖 564 个多样化场景和 86 种任务类型，支持丰富动作和环境组合，促进机器人通用操作技能学习 | 76k 条轨迹，350 小时交互数据 | 多任务学习，多机器人协同学习，跨场景泛化，多模态预训练 | $\pi_0$[43]，DiVLA[130]，DreamVLA[69]，HybridVLA[100]，NORA[129]，RDT[32]，SpatialVLA[70]，UniAct[131] |
| | Moblie ALOHA[132] | 大规模数据集，支持双臂移动操作，涵盖厨房、实验室等多场景下的复合任务学习，通过人类示教与自动采集数据 | 500 轨迹 | 多任务学习，导航学习，多模态预训练 | RDT[32] |
| | FrodoBots-2k | 大规模多模态数据集，遥控操作收集涵盖视频、GPS、IMU、音频与人类控制数据，覆盖全球 10 多座城市，支持移动机器人导航与感知研究 | 2k 小时交互数据 | 驾驶策略学习，跨场景泛化，多模态预训练 | HPT[34] |
| | OXE[33] | 大规模多机器人操作数据集，涵盖 22 种机器人的 527 种技能和 160k 项任务，提供标准化格式支持，促进跨形态经验迁移与通用策略学习 | 超过 1M 条轨迹 | 多任务学习，跨场景泛化，多模态预训练 | $\pi_0$[43]，CogACT[41]，CoT-VLA[67]，DiVLA[130]，GR00T N1[58]，HPT[34]，HybridVLA[100]，LAPA[50]，NORA[129]，RDT[32]，RoboVLMs[45]，SpatialVLA[70]，TriVLA[55]，UniAct[131]，UniVLA[66]，VPP[47] |
| | RDT-1B[32] | 大规模机器人操作数据集，涵盖单臂、双臂与移动机械臂等多种机器人形态 | 超过 1M 轨迹 | 多任务学习，跨形态泛化，跨场景泛化，多模态预训练 | RDT[32] |
| | RH20T[133] | 大规模多模态机器人操作数据集，包含 4 种主流机械臂、4 种夹爪和 3 种力传感器共 7 种机器人硬件配置组合，涵盖 147 种任务与 42 种技能 | 110k 序列，50M 帧 | 力觉感知融合，多形态技能泛化，多模态预训练 | RDT[32] |
| | RoboSet[134] | 大规模真实机器人操作数据集，专注于厨房环境，包含动觉示教与遥操示教的多视角轨迹及丰富场景变化 | 28.5k 轨迹 | 多任务学习，变化场景适应，多模态预训练 | RDT[32] |
| | RoboMIND[135] | 大规模机器人操作数据集，涵盖 479 种任务、96 种物体类别，38 种操作技能及多种机械臂与人形机器人，支持任务执行性能提升与失败案例分析 | 107k 成功轨迹，5k 失败轨迹 | 多任务学习，失败分析与自适应改进，多模态预训练 | HybridVLA[100] |
| | RT-1[22] | 大规模真实机器人数据集，包含 13 台机械臂上采集的带语言指令标注的视频，涵盖 700 多种任务，支持零样本泛化和复杂操作技能学习 | 130k 视频片段 | 多任务学习，跨场景泛化，多模态预训练 | Gen2Act[136]，GR-2[49]，RDT[32]，RT-1[22]，RT-2[11]，TraceVLA[127] |

演示轨迹（约 350 小时交互数据），覆盖 564 个真实场景和 86 项任务类型并且同步记录视觉、力觉、关节运动轨迹及环境数据。RDT-1B[32] 是一个多任务双臂操作机器人数据集，覆盖 300 多个挑战性任务，6k 多条轨迹，并整合了包括 RT-1 数据集在内的 5 个公开数据集，收集了 1M 条轨迹用于预训练策略。RoboMIND[135] 构建了包含 107k 条轨迹，479 种任务，96 种物体的数据集，通过采用统一的标准化收集协议，确保数据一致性和可靠性，减少变异性和噪声。智元机器人[65] 基于 AGIBot 机器人，使用人工遥操作的方式，采集了超过 1M 条轨迹，覆盖 217 种任务和 87 种技能。谷歌[33] 汇聚了 60 个目前公开的机器人数据集，构建了一个统一标准的数据集 OXE，涵盖了来自 22 个不同形态机器人的超过 1M 条轨迹，使用 RLDS 格式对数据进行了标准化，方便下载和使用。为了支持更多模态的 VLA 模型研究，上海交通大学[133] 构建了一个大规模多模态机器人操作数据集 RH20T，包含超过 110k 个密集接触的机器人操作序列，涵盖了 147 种不同的任务（包括 42 种技能），并且包含视觉、力、音频和动作信息，为多模态感知和环境交互提供了丰富的数据支持。鹏城实验室构建了 ARIO 数据集[137]，通过整合现有数据集和自采集数据集，构建了包含 2D 图像、3D 点云、声音、文本、触觉 5 种模态，3M 条轨迹的数据集，为不同形态的机器人数据提供标准化格式。

### 4.5　挑战与发展趋势

相比于 LLM 和 VLM 的训练数据规模，包含机器人动作的 VLA 训练数据集无论是从物体种类多样性、视角多样性、场景多样性以及数据规模等角度都相差甚远。当继承 LLM/VLM 权重后，使用规模相对较小的机器人数据集训练 VLA 时，会由于数据比例的巨大差异导致 VLA 模型过拟合到机器人轨迹数据中，丧失 LLM/VLM 预训练过程中获得的泛化能力。另一方面，现阶段的具身操作机器人难以通过人工操作的方式在真实的作业环境中长期积累大量的真实需求数据集，只能通过对相似的作业环境和任务进行模拟与仿真，导致难以发挥“数据飞轮”优势，实现真实场景数据的快速积累，从而阻碍具身操作实现真正的落地应用。考虑到真实数据的采集成本，打破数据屏障的关键可能在于如何充分利用数据金字塔中底层的数据，激发预训练 VLM 的潜力，是短期突破由于数据瓶颈导致的模型泛化能力的关键。另一方面，目前机器人轨迹数据中所包含的模态信息主要以视觉和机器人状态信息为主，种类相对单一，缺乏深度、触觉、听觉、力觉等多维信息，从而导致难以构建真正的多模态 VLA 模型，实现对于真实物理世界的全方位理解。高效的几何理解是提高 VLA 模型空间泛化和操作精度的关键之一，因此，将深度信息融合到 VLA 架构中是推动模型落地的关键因素之一。触觉和力觉在密集接触的场景下提供了额外的接触反馈信号，可以用于构建更安全和交互更友好的 VLA 模型。考虑到传感器的发展程度，融合力信号的 VLA 模型可能会更先一步进入工业场景，但从长远看，相比于力信号，触觉提供了更丰富信息，是 VLA 解决接触场景的理想信号来源。从长期来看，人类真正生活和作业场景的数据是驱动 VLA 落地应用的关键，无论是通过更灵活的数据采集设备或算法，还是通过机器人分阶段进入真实环境作业，大规模积累应用场景数据，转动具身操作的“数据飞轮”，提升机器人轨迹数据的多样性与规模，是突破当前 VLA 瓶颈的关键之一。通过在仓储物流或无人商超等典型相对封闭场景下部署 VLA 模型，在作业场景中积累场景数据和人类接管数据，从而迭代模型以解决更多的场景，向开放场景下逐渐扩展。

## 5　VLA 预训练

VLA 模型预训练方法与数据集息息相关。监督学习作为目前 VLA 预训练的主要手段，不同的数据组合方式和训练方法对其性能的影响是非常重要的。通过对当前 VLA 预训练过程进行梳理，根据训练过程中使用的数据和训练方式，本文将目前 VLA 预训练方法分为 4 种：单一领域数据训练、跨域数据分阶段训练、跨域数据联合训练和思维链增强。图8展示了不同的训练方法和使用的数据集类型。

### 5.1　单一领域数据训练

不同于大语言模型无监督的预训练范式，VLA 模型的预训练通常使用大量的机器人轨迹数据监督训练完成。Octo[31] 使用大规模公开机器人数据集 OXE 进行预训练，然后再使用当前任务的数据集进行微调。RDT[32] 构建了大规模双臂数据，先在公开的机器人数据集上预训练，然后再在收集的双臂操作数据集上微调，获得优异的双臂操作性能。相比于 LLM 预训练使用的海量互联网数据，尽管目前已经公开了多种不同形态的机械臂操作数据集，但其数量以及多样性相比于大模型预训练数据集相差甚远。从而导致该训练模式下所得到的 VLA 模型的视觉泛化能力和任务泛化能力比较差。

### 5.2　跨域数据分阶段训练

考虑到目前已有的预训练大语言模型和视觉语言模型可以提供优异的文本理解和环境理解能力，因此，OpenVLA[40] 通过继承预训练的 LLM 模型 LLaMA 的权重，并在此基础上对齐机器人操作任

务的视觉编码器和动作序列。由于 LLaMA 在预训练中仅使用了文本数据，无法获得良好的视觉理解先验。相比于继承 LLM 权重，直接继承 VLM 权重可以使得 VLA 获得更强的文本与视觉理解先验。RT-2[11] 虽然没有直接继承 VLM 权重，但是在预训练过程中使用了大量的互联网图文数据，使其获得 VLM 能力，然后再通过使用机器人操作轨迹数据进行微调，对齐机器人操作能力，也可以获得良好的泛化性能。

虽然视觉语言模型提供了优异的视觉泛化能力，VLM 训练的数据通常是静态的图像-文本数据对。而机器人操作是一个序列决策任务，通常需要模型捕捉时序运动信息。视频作为一种包含时间和空间信息的媒介数据，也被用于辅助训练 VLA 模型。UniPi[46] 使用文本条件视频生成模型可以作为机器人通用规划策略，用于合成多样化的行为，通过少量机器人轨迹数据训练的逆动力学模型，可以直接生成机器人策略。GR-1[48] 将视频预测与动作生成融合在一个模型中，先使用大规模视频数据进行预训练，然后再使用机器人轨迹数据进行微调。GR-2[49] 在 GR-1 的基础上，通过引入 VQ-GAN[138] 增强模型处理多视角图像能力，并使用了更大规模的文本-视频数据进行预训练，进一步提升了模型泛化能力。考虑到视频数据中缺乏机器人动作标签，LAPA[50] 通过视频中的连续两帧图像，结合 VQ-VAE 学习离散潜在动作空间表示。然后通过使用人类操作视频数据或机器人操作视频数据，在预训练 VLM 权重的基础上微调使其具备预测离散潜在动作的能力，最后使用少量机器人数据构建潜在动作到真实机器人之间的映射。GO-1[65] 结合 VQ-VAE 和视觉图像重建，构建了潜在动作空间作为文本-图像-视频数据到机器人动作数据的桥梁。通过使用机器人轨迹数据进行微调，构建潜在动作空间到真实机器人动作映射，从而提升了 VLA 模型对于异构数据的扩展能力。UniVLA[66] 通过将图像变化分解为动作所导致的变化和其他外部因素导致的变化两部分，构建以任务为中心的潜在动作空间学习方法。通过使用视频数据预训练，结合动作解码器输出机器人真实动作，可以实现对纹理、光照等因素的更强抗干扰能力。由于人类活动数据与机器人操作数据行为差异比较大，这种隐式动作空间不一定能够非常高效地桥接人类动作和机器人动作，EgoVLA[111] 和 Being-H0[112] 提出使用带有人类手部动作标记的第一人称视角下人类活动数据对 VLA 模型进行预训练，然后再使用少量真实机器人数据进行微调，实现技能的迁移。

### 5.3 跨域数据联合训练

虽然使用大规模互联网数据和跨域数据对 VLA 模型进行预训练可以大幅提升模型对于场景的泛化能力，但是由于机器人轨迹数据与预训练数据分布差异比较大，从而导致模型在使用机器人轨迹数据微调时，会遗忘预训练获得的能力，产生“虚假遗忘”现象。为了缓解这个问题，很多研究工作尝试通过跨域数据联合训练的方式，强制模型在不同任务时共享参数，从而保持已获得能力。早期 GATO[26] 融合了图像描述、对话聊天、游戏控制、机器人控制等 604 个不同任务数据集训练一个通用策略模型，可以由同一个模型完成多种不同的任务，展示了 Transformer 结构的跨模态学习和迁移能力。UP-VLA[64] 在 VLA 训练过程中结合了多模态理解和未来预测两个任务，增强模型对于物理动态和空间细节的理解能力，在提升动作预测能力的同时保持了多模态理解能力。近期的研究工作表明在训练过程中整合多种异构数据源，包括多模态理解任务数据、目标检测与定位任务数据以及跨形态机器人数据，可以大幅提升 VLA 模型的泛化能力[79,101,139]。跨域数据联合训练通过强迫不同任务共享参数，虽然能够提升模型泛化能力，但是也会导致不同任务之间的参数竞争。ChatVLA[51] 针对对话数据，多模态理解数据和机器人轨迹数据，设计了基于混合专家架构的主干网络，将机器人控制与环境理解隔离，以防止不同任务之间相互干扰。但这种方式也一定程度上阻碍了不同任务之间的知识共享。为了更好地解决图像-文本数据与机器人轨迹数据在联合训练时共享参数竞争的问题，ChatVLA-2[95] 在前述工作的基础上，使用动态混合专家架构代替确定性混合专家架构，可以保持 LLM 的完整架构，确保基础知识不受干扰，并且可以选择性激活以适应特定任务。从目前的发展过程来看，跨域联合训练，即在模型训练过程中同时利用来自不同领域或模态的数据，使模型能够学习多样化特征从而提升其泛化能力，在激发预训练 VLM 知识，提升 VLA 泛化能力方面具有巨大潜力。如何选择合理的联合训练数据，以及构建合适的模型架构可能是构建下一代 VLA 模型的关键。

### 5.4 思维链增强

跨域数据联合训练是通过强制不同的任务贡献参数增强模型的泛化能力，这种方式有可能会因为任务之间的差异性导致参数竞争现象。在大语言模型领域，通过激发模型思维链的方式，激活模型与当前任务相关的基础能力，从而提高模型性能。这种方式也被借鉴到 VLA 模型领域，通过让 VLA 模型理解场景和任务，缓解观测到动作的“肌肉记忆”导

致的泛化能力比较差问题。ECoT[62] 强制模型在生成动作前先进行多步推理。其中多步推理包括任务分解，子任务规划等高层次推理以及物体和末端执行器检测等视觉感知推理。EMMA-X[63] 在 ECoT 的基础上，通过使用 VLM 对运动过程，执行轨迹以及空间状态进行标注，生成了更精细的推理过程。RAD[140] 利用语言推理作为桥梁，能够同时利用包含思维链的机器人轨迹数据和无动作的人类视频数据，解决了因缺乏动作标签而导致人类视频无法用于 VLA 模型训练的问题。ReasonManip[141] 将机器人操作重新表述为多轮数学推导问题，通过基于正交轴空间表示方法，允许 VLM 通过语言直接推理中间目标和末端执行器姿态。Inspire[142] 通过让模型表述操作与物体之间的空间关系，构建空间推理问题，让 VLA 模型先进行空间推理，然后再输出动作，从而提高模型的泛化能力。DiVLA[130] 通过推理注入模块，让 VLA 模型在生成动作前先进行思考和推理，并把推理结果注入到动作解码网络中，将“思考”与“行动”更紧密地绑定在一起。ChatVLA-2[95] 使用 GPT-4o 或预定义目标构建多个数据集的推理过程。对于机器人数据，构建了数学匹配任务和简单物体放置任务的推理数据。与 DiVLA 类似，该工作也将推理结果注入到动作解码过程中，提高推理与执行的一致性。

推理过程虽然增强了模型对于环境和任务的理解能力，但是也增强了模型的输出长度和时间延迟，这对于机器人的实时控制是个挑战。OneTwoVLA[143] 提出了自适应推理机制，通过引入 2 个特殊的令牌，让模型只在关键时刻思考，其他时刻则是根据已有的思考结果直接行动，从而避免每一步都需要进行思考导致，减小输出时间延迟。ECoT Lite[144] 介绍了一种推理随机丢弃的方式，通过在训练过程中随机丢弃部分或全部推理步骤，让模型学会在有无推理的情况下都能预测动作，在测试时可以灵活选择是否生成推理，实现速度和性能的平衡。Fast ECoT[128] 提出了一种推理时加速方法，通过缓存和重用中间推理步骤将推理和动作生成解耦，允许在后台更新推理的同时发出高频控制信号，实现了部分并行的思维链生成，显著提高了响应速度。与跨域数据联合训练方式不同的是，思维链增强希望通过思考过程激发模型自身能力，并在 VLA 训练过程中始终保持该能力，而不是通过借助外部数据强化其理解能力，但是目前该方法也面临一些挑战: 首先，构造何种思维链数据仍然是一个开放性问题，不同的思维链数据对应于不同的能力，哪种思考模式最适合 VLA 模型训练还有待探索；其次，目前通过监督学习将外部思维链数据内化是主要的手段，是否有其他的解决方法能够自发地激发 VLM 本身思考能力而不是通过大量外部数据的方式也是一个值得探索的方向。

### 5.5 挑战与发展趋势

VLA 预训练的目标是赋予 VLA 模型理解机器人控制任务的能力，目前主要以基于机器人数据的模仿学习范式为主，在预训练 LLM/VLM 权重基础上，实现对机器人控制动作的映射。但是由于机器人轨迹数据的多样性和规模相对较小，导致模型的泛化能力和指令跟随能力比较差。除了扩大机器人轨迹数据，使用低成本的图文数据，与机器人轨迹一起进行联合训练，已经被验证是一种非常高效且有前景的预训练模式。由于预训练机器人数据通常是有多种不同形态的机器人采集，不同形态机器人在相同任务下控制动作的差异性通常会在预训练过程中被平滑，从而导致预训练模型对形态的泛化能力比较差。尽管目前已经有工作探索如何在预训练过程中使用机器人形态信息，进一步挖掘形态信息在预训练过程中的作用，仍然是跨形态通用 VLA 需要解决的问题之一。此外，由于力感知状态的缺失，现阶段 VLA 模型难以理解带修饰的交互动作，例如“轻轻地”，“用力地”等，如何在预训练过程中，将这种多模态信息对齐到 VLA 模型中，仍然是有待探索的方向。

## 6 VLA 后训练

考虑到 VLA 模型目标是构建面向开放环境的机器人通用策略模型，预训练阶段会使用大规模真实机器人采集的公开数据集，涵盖丰富的任务、环境、机器人形态和传感器设置等，为 VLA 模型泛化能力奠定基础。当把这种预训练的模型部署到真实的机器人上时，通常会面临性能很差甚至不工作的问题。这种问题与大模型中的“幻觉”问题比较类似——通常是编造一些不存在的“事实”来回答问题。在机器人中的表现在于无法拒绝未在训练过程中出现的场景，从而产生不可靠行为。主要原因在于部署机器人与数据中机器人本体不一致，以及数据中任务和环境与真实不一致引发的数据分布不匹配，从而导致模型部署性能下降。因此需要根据目标机器人和任务采集少量数据进行后训练。按照大模型后训练分类方法[145]，本文将目前的 VLA 后训练方法也分为 3 类：监督微调、强化微调和推理扩展。详细的方法介绍见表3。

### 6.1 监督微调

监督微调是目前 VLA 模型后训练过程中最广泛使用的方法，通过采集少量任务相关的数据，使用监督学习的方法，在预训练 VLA 模型权重的基础上继续微调，从而使得模型适应当前的机器人和

表 3: VLA 后训练方法汇总

Tab. 3: Summary of VLA Post-training Methods.

| 类别 | 相关工作 | 主要贡献/优势 | 缺陷 | 适用场景 | 发表 | 时间 |
| --- | --- | --- | --- | --- | --- | --- |
| 监督微调 | $\pi_0$ [43] | 提出基于流匹配的动作解码器，在高质量真实机器人数据上通过监督微调，有效提升模型在复杂、长程任务中的执行稳定性与成功率 | 对专家数据的质量与覆盖度较为敏感，分布外泛化与跨场景迁移受限，需要进行域内微调 | 适用于长时序任务及不便进行在线探索的真实部署环境 | RSS | 2025 |
| | GO-1 [65] | 在海量互联网异构视频与真实机器人数据上预训练，并结合 MoE 结构，仅需少量真实数据的监督微调即可实现快速适应新任务与新场景，显著降低数据需求 | 预训练数据噪声与域间分布差异产生错误对齐；MoE 带来训练与推理的系统复杂度与算力开销，缺乏稀有技能与密集接触场景真机数据 | 数据相对稀缺但任务多样、需快速落地的新场景迁移 | IROS | 2025 |
| | GR00T N1 [58] | 在多样化人形机器人感知与控制数据上预训练，结合快速反射与规划推理的双系统架构，后训练后既具备高反应速度，又能进行复杂任务规划，在多种场景中展现出鲁棒的人形机器人控制能力 | 依赖大规模高质量人形数据与复杂系统集成，训练与推理的算力/工程成本高；密集接触工况或强约束环境中需要额外安全策略与调参 | 需要同时具备灵敏即时反应和高层规划的人形机器人应用，例如服务场景、多步骤装配、移动操作与协作任务 | arXiv | 2025 |
| | GR-1 [48] | 在 400k 条跨形态机器人数据上进行模仿预训练，构建首个开放的多任务、多机器人形态统一策略基线，在少量目标机器人真实数据上通过监督微调进行后训练，实现了“单模型多机器人”控制的可行性，并显著降低多形态适配成本 | 依赖大规模异构数据的质量与覆盖度；对极端工况/特定形态的细粒度控制可能仍需额外调参，且通才策略在个别边缘任务上可能不如专才策略 | 多形态、多任务的统一部署与快速落地，以及对维护成本敏感、需快速适配新形态的应用 | ICLR | 2024 |
| | GR-2 [49] | 在 GR-1 框架基础上加入视觉-语言-动作三模态对齐，并引入更大规模的互联网视频自监督数据进行预训练，在少量目标机器人真实数据上通过监督微调进行后训练，进一步提升了复杂指令理解和跨场景任务执行的泛化能力 | 依赖互联网视频与多模态对齐质量，可能受噪声与域间偏移影响；模型规模与训练/对齐流程复杂，密集接触任务的安全性与精细控制仍需额外工程与人为监督 | 真机数据稀缺但可获取大量弱标视频的应用；需快速适配新环境/新任务的指令驱动型操作；跨机器人形态迁移与多步骤长程任务执行 | arXiv | 2024 |
| | GR-3 [139] | 采用跨域数据联合训练，将预训练数据规模扩展至百万级，重点强化语言指令理解与零样本跨环境泛化能力，在少量目标机器人真实数据上通过监督微调进行后训练，显著提升了在未见任务与新环境中的执行稳定性与成功率 | 高度依赖大规模异构数据的质量与对齐，数据清洗与标注成本高；训练与部署的系统/算力开销较大，密集接触与高精度控制仍可能需要额外专用微调策略 | 数据分布多样、需频繁迁移的新环境任务；指令驱动的长程多步骤操作 | arXiv | 2025 |
| | Helix [57] | 首个人形 VLA，在 Figure 人形的大规模感知与控制数据上进行预训练，可对整个人形上半身输出高频连续控制，在少量目标任务的真实机器人数据上通过监督微调进行后训练，实现精确且稳定的上肢协调控制 | 训练数据仅限于当前机器人形态，迁移到异构人形可能需额外标定与微调；高频连续控制带来训练与推理算力/实时性压力，密集接触场景仍需安全机制与细致调参 | 需要精细上肢操作与稳定协同的人形应用，例如装配、工具使用、开关/旋钮操作与服务场景 | arXiv | 2025 |
| | HPT [34] | 分层提示微调，将 LLM 生成的文字描述拆分为“层次化结构”与“语义文本”两路提示并同步学习，在多任务机器人数据上预训练后，用少量目标任务的真实数据监督微调后训练，保持语言理解能力同时显著提升任务执行的稳定性与泛化性 | 需要高质量的层次化指令生成与标注，分层提示设计与超参较多、工程复杂度偏高；对密集接触/高频闭环控制仍可能需额外控制器或安全机制配合 | 语义复杂、步骤明确的多步骤操作，例如装配、烹饪式流程、服务机器人任务 | NeurIPS | 2024 |
| | Magma [125] | 微软提出的多模态基础模型，可同时感知视觉与语言并输出动作，在少量目标机器人真实数据上通过监督微调进行后训练，实现了从数字智能体到实体机器人的高效迁移，在真实环境任务中表现出稳定的感知-行动能力 | 预训练与对齐流程复杂，对多模态同步标注与时间对齐敏感；推理开销与系统集成成本较高，密集接触或高精度操作仍需专才策略与安全约束 | 需要从仿真/视频智能体快速落地到真实机器人、任务多样且数据相对有限的应用，例如服务机器人、仓储与装配等真实场景的多任务部署 | CVPR | 2025 |
| | Octo [31] | 首个完全开源的通用 Transformer Diffusion 架构，在少量目标机器人真实数据上通过监督微调进行后训练，实现了跨机器人形态的快速适配与任务迁移，在多样任务中保持高执行性能 | 扩散式动作生成推理开销较大，对密集接触/高精度操作仍可能需要专用微调方法与安全机制；性能对专家数据分布与对齐质量较为敏感 | 跨形态迁移与多任务统一基线搭建、研究与工业落地的可复现方案，以及以少量目标数据完成快速适配的真实部署 | CoRL | 2023 |
| | OpenVLA [40] | 7B 参数开源 VLA，在少量目标机器人真实数据上通过监督微调后训练，内置多机器人形态的适配能力，从而实现了高效迁移，显著降低了跨形态适配成本并且可以保持任务的执行性能 | 模型规模与推理开销较大，对实时性与边缘设备部署有压力；在密集接触/高精度任务上仍依赖高质量对齐与额外标定与安全机制 | 作为跨形态统一基线与研究/工业的可复现方案，在数据有限的场景进行快速适配与任务迁移，可在多机器人形态间共享策略 | CoRL | 2024 |
| | RDT [32] | 首个双臂操作扩散基础模型，在稀缺数据场景下能够生成多模态动作分布，在少量目标任务的真实双臂机器人数据上通过监督微调进行后训练，显著提升了在复杂协作操作任务中的稳定性与成功率 | 扩散生成带来推理时延与算力开销，对实时高频控制有压力；对精细力控与密集接触场景仍需额外传感/控制层或专用微调方法，且对演示对齐质量较敏感 | 双臂协作的装配、整理、搬运与工具协同等任务，数据有限但需高稳定性的工业/服务场景署 | ICLR | 2025 |
| | RoboFlamingo [38] | 以 OpenFlamingo 作为视觉-语言底座，在少量目标机器人真实演示数据上通过监督微调进行后训练，在多任务指令条件下显著提升了执行成功率与泛化能力 | 性能对指令—动作对齐质量与专家轨迹丰富度敏感；在密集接触、强时延约束或高频闭环控制任务上仍受限，需要额外控制器与安全机制 | 指令驱动的多任务桌面操作与服务场景、数据有限但可快速收集少量演示的部署 | ICLR | 2024 |
| | RT-2 [11] | 使用互联网数据和机器人轨迹数据预训练实现端到端机器人控制，在少量目标机器人的真实演示数据上通过动作映射进行后训练，赋予模型“语义推理”能力，并在未见场景下显著提升任务成功率 | 依赖大规模跨域数据的对齐与清洗；对密集接触/高精度控制需额外控制器与安全机制支持，推理与部署成本较高；操作泛化能比较差 | 指令驱动、语义复杂且环境多变的服务/家居/仓储任务 | CoRL | 2023 |
| | UniVLA [66] | 将视觉、语言与动作离散化为统一令牌序列，并用单一自回归 Transformer 进行统一建模，在结合世界模型预训练后，在少量目标机器人真实数据上通过监督微调进行后训练，显著提升了长时序任务的迁移能力与执行稳定性 | 离散化与自回归解码在高频控制下存在时延与信息损失风险；世界模型预训练和对齐流程复杂，对数据质量与时间对齐敏感 | 需要长程规划与多步骤执行的指令驱动任务、跨形态/跨场景迁移场景 | RSS | 2025 |

| 类别 | 相关工作 | 主要贡献/优势 | 缺陷 | 适用场景 | 发表 | 时间 |
|---|---|---|---|---|---|---|
| 监督微调 | VPP [47] | 将视频扩散模型的未来表征嵌入策略网络，以隐式方式学习逆动力学，在少量目标任务的真实机器人数据上通过监督微调进行后训练，显著提升了长时预测下的控制稳定性与样本效率 | 依赖视频扩散模型的质量与时序对齐，训练/推理开销较高；在密集接触与精细力控任务中可能存在动力学失配，需额外控制器或微调 | 长程、多步骤、需要前瞻规划的操作，例如装配、整理、导航取放 | ICML | 2025 |
| 强化微调 | ConRFT [146] | 在预训练 VLA 的基础上，采用一致性策略并结合人为干预，通过在线强化学习在真实机器人上进行强化微调后训练；仅需 45–90 分钟即可将任务成功率提升至 96% | 需要在线交互与人类干预，工程与系统集成复杂度较高；对一致性目标/超参较敏感，迁移到密集接触或新设备时仍需额外调参与校准 | 真实机器人上的快速任务适配与性能冲刺，尤其是密集接触、风险较高且需稳定性的工业/服务场景 | RSS | 2025 |
| | GRAPE [147] | 利用 VLM 将复杂任务分解为子目标并生成轨迹级偏好奖励，在真实机器人交互数据上通过直接偏好优化（DPO）进行后训练，无需额外人工标注即可提升任务成功率与与人类偏好的一致性 | 偏好可信度受 VLM 评估与分解粒度影响，可能引入噪声或阶段性误导；对长程依赖与接触密集场景仍需精心设计阶段/约束，训练稳定性对数据分布较敏感 | 难以手工设计奖励、但能获取交互数据的真实部署；需要对齐人类偏好（安全、舒适、效率等）的服务/家居/协作操作，以及多目标权衡的长程任务 | ICRA | 2024 |
| | iRe-VLA [148] | 提出迭代式 RL-SFT 环（内环强化微调，外环监督微调），在真实与仿真任务数据上仅更新轻量动作解码器进行后训练，实现了高样本效率的稳定收敛，并在多任务中保持良好的泛化性能 | 依赖在线交互与环路调度，例如 PPO 超参、数据混合比例，对安全与复位机制有要求；骨干冻结限制了感知侧的进一步提 | 真机/仿真均可交互、真实数据有限但需快速适配与稳健收敛的多任务部署 | ICRA | 2025 |
| | PARL [149] | 在预训练 VLA 的基础上，通过 Q 函数迭代优化动作，并以模仿学习方式学习这些优化后的动作，在真实机器人数据上进行后训练，稳定提升了模型的任务执行性能 | 质量高度依赖 Q 函数的准确性与覆盖度；若 Q 学习出现过估计，将把错误信号蒸馏进策略；在线阶段仍需一定交互与工程调参（候选采样、优化步数等） | 具备一定离线数据或 Q 容易训练的真实/仿真操控任务；希望在不改动大模型骨干的前提下，以低风险方式持续提升任务执行性能的工业与服务场景 | ICLR | 2024 |
| | Policy Decorator [15 | 将大型离线模仿策略作为基础策略，在真实机器人数据上在线叠加可学习残差控制器，结合受控探索与信任域优化进行强化微调后训练，实现对下层策略模型不可知、稳定且高效的性能提升 | 性能上限仍受基础策略能力与误差耦合制约；残差与主策略的协同需要细致的权重/约束设计，可能引入额外超参调试成本 | 已有成熟模仿策略部署、需在真实环境中快速提效且不希望改动主干的部署 | ICLR | 2025 |
| | ReinboT [151] | 将强化学习的累计回报目标显式融入 VLA 损失函数，在真实机器人数据上通过稳定的训练流程进行训练，提升了任务执行性能并保持训练收敛的稳定性 | 价值学习容易受到奖励设计/标注与时序的影响；引入回报条件与价值分支增加系统与算力开销 | 具备一定真实交互或离线回放数据、希望在不大改骨干的前提下系统性提效的多任务部署 | ICML | 2025 |
| | RIPT-VLA [152] | 在 1-demo 监督微调起点上，通过交互式强化微调并结合 RLOO 确保梯度稳定，在仿真环境中完成全部实验，成功率可提升至 97%，验证了在极少示例条件下的高效任务学习能力 | 主要在仿真中验证，真实部署的感知噪声与复位/安全成本未充分评估；对二元成功信号与采样分组策略敏感，探索策略易陷入局部最优 | 数据极度稀缺但可进行大量模拟交互的场景；具有明确成败判据的短/中程操控任务 | CVPR | 2025 |
| | RLDG [153] | 在真实机器人上通过强化学习生成高质量的自监督轨迹，并将这些“内生数据”蒸馏回大模型，无需额外人类示范即可迭代提升模型性能，实现真机环境下的高效自我改进 | 依赖稳健的在线 RL 基础设施与安全复位机制，训练易受奖励设计/不稳定性影响；真机交互与算力成本仍不低，且策略采样数据可能带来偏置与遗忘风险，需要精心的数据筛选与蒸馏配方 | 具备可自动判定成败/奖励的操作任务与可用机器人集群的场景；希望以极少人类示范实现长期、持续迭代学习的真实部署 | RSS | 2025 |
| | TGRPO [154] | 在 GRPO 基础上进行拓展，同时利用时间步级与轨迹级奖励对模型进行微调，全部实验均在仿真环境中完成，显著提升了连续控制任务中的动作质量与长时序一致性 | 目前主要在仿真验证，真实环境中的感知噪声、延迟与安全约束尚未系统评估；组内采样与优势归一化等超参较敏感，奖励/成功判据设计不当可能削弱收益 | 需要长程依赖与连续控制的多步骤操控，例如装配、抓取—放置序列 | arXiv | 2025 |
| | VLA-RL [155] | 在预训练 VLA 的基础上，结合 PPO 与 RPRM [155] 进行微调，并利用并行仿真环境加速训练，全部实验均在仿真中完成，实现了更快的收敛速度与更高的任务执行性能 | 基于仿真验证，真实环境的传感噪声、延迟与安全约束尚未充分评估；超参与奖励设计敏感，仿真到现实存在潜在性能落差 | 需要大规模离线/并行仿真进行策略筛选与组合搜索的多任务操控 | arXiv | 2025 |
| 推理扩展 | FOREWARN [156] | 利用 VLM 的轨迹评估能力，对 VLA 生成的多个候选动作规划进行评估筛选，在真实机器人数据上完成全部训练与验证，避免了奖励函数设计和值函数学习的需求，并提升了任务执行的稳定性与成功率 | 依赖 VLM 评估的可靠性与无偏性，易受分布外场景与部分可观测性的影响；多候选采样与评估带来计算与时延开销 | 难以手工设计奖励的真实部署任务、需要快速稳健提效而不改动训练流程的场景，以及评估器判断可靠性高下的工业与服务机器人应用 | RSS | 2025 |
| | Hume [61] | 训练分层控制系统的 S2 系统通过预测值函数以评估生成动作序列的质量，在仿真与真实机器人数据上进行训练与验证，提升了动作决策的可靠性与任务执行性能 | 分层架构与价值评估带来系统与推理复杂度、实时性开销；价值学习对数据与超参敏感 | 需要深度推理与高频控制并存的复杂长程任务，例如装配、工具使用与多步骤服务操作 | arXiv | 2025 |
| | ITPS [157] | 在动作生成过程中允许人类通过交互方式输入轨迹偏好，并通过引导扩散过程生成期望的动作序列，在仿真与真实机器人数据上进行训练与验证，提升了模型对人类意图的响应能力与任务执行的可控性 | 需在线人机交互与额外推理开销；偏好可行度低或引导强度不当可能导致过度约束或目标漂移，对实时性与稳定性提出更高要求 | 需要按用户偏好动态定制行为的服务/协作任务、对安全与舒适有要求的真实部署 | ICRA | 2025 |
| | RoboMonkey [158] | 提出一种“采样-验证”的推理期扩展框架，并验证了动作误差与生成样本数量之间近似符合幂律关系，在仿真与真实机器人数据上进行评估，有效降低了推理过程中的动作错误率并提升了任务成功率 | 依赖高质量验证器与评估信号，分布外场景可能失效；多候选采样与验证增加推理时延与算力开销，在实时高频控制与密集接触任务中需谨慎权衡 | 允许较大推理预算以换取稳健性的部署，例如离线规划、低速高精度操作、关键任务执行前的安全校验 | RSS | 2025 |
| | V-GPS [159] | 针对同一指令在 VLA 上并行采样多条候选动作轨迹，并利用实时视觉估计进行评分，选择并平滑最优轨迹后下发控制，在仿真与真实机器人数据上进行验证，有效提升了安全性与成功率 | 依赖在线视觉感知与评分器可靠性，多候选采样与评估增加推理时延与算力开销；在密集接触操作中，视觉滞后与估计误差导致评分偏差 | 对安全与稳健性要求高、允许一定推理预算的真实部署，例如抓取与装配、拥挤/狭窄环境操作 | CoRL | 2024 |

任务[31-32,38,40,43]。在监督微调中，数据质量对于微调后的模型性能非常重要。目前监督微调使用的数据通常是通过遥操作采集的。操作员通过操作数据采集设备控制机器人，采集机器人轨迹数据。目前的数据采集方案可以大致分为 3 大类：第 1 类采用主-从臂的方式，即构建包含两套相同或相似的机械臂，其中一套作为主动臂，一套作为跟随臂。操作员通过握持拖动主动臂，然后将主动臂的状态映射到跟随臂上，从而控制跟随臂完成任务。第 2 类是采用虚拟动作映射的方式，通过动作捕捉设备或人体关键点检测模型，捕捉人类手臂的运动状态，然后通过构建人类手臂与机械臂的动作映射，从而控制机器人完成任务。第 3 类是采用末端映射的方式，不同于前两类操作员直接操作或映射关节，这类方法只需要控制机械臂末端，采集末端数据。斯坦福大学的研究人员设计了一种不依赖机械臂的数据采集方案[160]，大大缩减了数据采集成本，提高了数据采集灵活性和效率。监督微调是通过让模型模仿操作员行为的方式，赋予模型完成任务的能力。虽然监督微调简单高效，训练稳定，但是其对数据质量要求非常高。因此，监督微调适用于操作精度不高、数据获取容易以及任务多样性高的场景。由于在数据采集过程不同操作员的行为倾向不一致或同一个操作员时空上的行为不一致导致数据的一致性难以保障，从而限制了模型的性能上限[153]。此外，作为模仿学习的一种，监督微调同样面临连续决策过程中的复合误差问题，导致该类方法在长时间连续作业任务上面临巨大挑战。

### 6.2 强化微调

从大语言模型的发展路径来看，强化学习在提升大语言模型能力方面具有巨大潜力，尤其是近期以 OpenAI o1 和 DeepSeek R1[161] 为代表的推理大模型，展现出强化学习在后训练方面卓越的能力。此外，强化学习在机器人控制领域也不断突破性能上限，因此，使用强化学习做 VLA 后训练似乎是一个合理的选择。但需要注意到的是，相比于大语言模型的强化学习后训练，由于 VLA 需要与环境交互才能产生数据，使用强化学习对 VLA 进行后训练面临更多的挑战。目前使用强化学习实现 VLA 后训练的方法可以大致分为 2 类。其中一类是使用强化学习训练小模型或模型部分参数，并使用训练后的策略收集数据，然后在该数据的基础上进行监督微调[153,149,148]。由于强化学习策略采集的数据具有更好的一致性，因此，这类方法可以获得相比于使用遥操作数据监督微调更好的性能。

另外一类方法是使用强化学习直接微调 VLA 模型权重。ConRFT[146] 提出了一个两阶段 VLA 强化微调方法，首先通过采集的数据对预训练的策略模型进行微调，并对强化学习所需要的值函数进行预训练，然后再通过人在回路的强化学习实现在线快速微调。针对在线微调过程中稀疏奖励导致的训练不稳定问题，RIPT-VLA[152] 使用去一法（Removed-Leave-One-Out，RLOO）估计近端策略优化（Proximal Policy Optimization，PPO）训练所使用的值函数，并结合动态拒绝采样稳定训练过程梯度。VLA-RL[155] 提出了轨迹级强化学习微调范式微调自回归模型，使用微调后的 VLM 作为过程奖励模型，缓解稀疏奖励问题。GRAPE[147] 利用在线交互过程中产生的成功与失败轨迹隐式建模奖励函数，提出轨迹级的偏好对齐方法，缓解 VLA 模型泛化能力不足的问题。受到组相对策略优化（Group Relative Policy Optimization, GRPO）的启发，轨迹级组相对策略优化（Trajectory Group Relative Policy Optimization, TGRPO）[154]融合了轨迹相对优势和步骤相对优势两个层次的优势信号，改进了 GRPO 的组级优势估计，使算法更适合 VLA 的在线强化学习训练。ReinboT[151] 将强化学习最大化累积回报的核心思想融入端到端 VLA 架构中，使用回报作为当前策略的输入条件，从而利用不同质量的轨迹，可以从混合质量数据中学习鲁棒策略。对于如何选择强化学习后训练方法，有研究工作构建了涵盖视觉感知、语义理解与执行稳健性 3 条评测维度的综合基准，揭示传统监督微调在遇到分布漂移时易积累误差、导致泛化受限的问题，并系统地评估了 PPO，DPO 和 GRPO 等强化学习在提升 VLA 模型泛化能力方面的独特价值[162]。

相比于监督微调，虽然强化微调能够获得更好的性能上限，但是其训练难度更高，训练稳定性比较差。并且对于一些依赖仿真环境的方法，要求仿真环境与真实环境非常相似，才能保障训练后的模型可以用到真实的场景中。因此，目前的强化微调技术适用于操作精度要求比较高的特定场景。

### 6.3 推理扩展

无论是监督微调或者强化微调，都需要额外的机器人数据和大量的计算资源。与 LLM/VLM 后训练方法相似，目前也发展出了一些针对 VLA 模型的推理时扩展的后训练方法。加州大学伯克利分校提出的 V-GPS[159] 通过在 VLA 模型推理时采样多个动作，并使用离线强化学习预训练的值函数对多个动作进行评估，并选择最优的动作执行。这种方法可以缓解实际部署环境与训练环境存在的分布差异，并且不需要访问或修改原始策略的权重，可与黑盒模型配合使用，但是仍然需要大量的数据来训练一个稳定的值函数。FOREWARN[156] 则通

过发挥 VLM 的轨迹评估能力，通过对 VLA 生成的多个候选动作规划进行评估，从而避免了奖励函数设计和值函数的学习，可以获得更稳定的评估结果。ITPS[157] 则是在动作生成过程中允许人类通过交互的方式输入轨迹偏好，通过引导扩散过程从而生成期望的动作。与 LLM/VLM 中类似，推理时扩展面临同样的问题：其输出的策略性能严重依赖于预训练 VLA 的性能，并且输出结果不稳定。RoboMonkey[158] 使用了一种“采样-验证”的推理时扩展框架，并且该工作证明了动作误差与生成样本数量之间遵循近似幂律关系，为机器人策略的测试时扩展（Test-Time Scaling）提供了理论基础，进一步提升了 VLA 模型推理时的性能。Hume[61] 在训练分层系统中的系统 S2 时借鉴强化学习值函数思路，使其能够预测值函数以评估生成动作序列的好坏。在推理时，通过多次采样后选择最优轨迹，发送给系统 S1 跟随该轨迹。

虽然推理扩展避免了离线的数据微调过程，但是其要求在推理过程中计算多条轨迹并评估，从而导致其推理时间代价更高。因此，推理扩展很难满足高频的实时控制要求。它的优势在于可以发挥动作或轨迹评估模型的泛化能力，处理推理过程中的不确定性。

### 6.4 挑战与发展趋势

与传统大模型类似，目前 VLA 训练过程也分为预训练和后训练 2 个阶段。通过后训练过程，VLA 模型可以快速对齐到当前的机器人形态和任务中，并且能够在新的场景中持续提升操作性能。目前大部分的 VLA 模型的后训练是通过采集大量遥操作数据，并由监督学习训练完成。但是由于人类遥操作采集数据的不一致性和监督学习在连续决策任务上的复合误差问题，导致该路径面临显著的性能瓶颈。受到强化学习在大语言模型后训练中进展的启发，有一些研究工作开始探索使用强化学习进行 VLA 模型的后训练，并且验证了强化学习不仅可以提高模型操作性能，一定程度上也能提高模型的泛化能力。目前大部分工作集中在仿真环境中，需要依赖精确的奖励函数、可大规模并行的环境、上百万次的环境交互等理想条件。但是在实际的操作环境中，这些条件往往很难满足。首先，在真实环境中获得精确的奖励函数非常困难，简单的分类器生成二值化奖励函数对于单一任务非常高效，但是很难扩展到开放环境任务中。虽然利用 VLM 可以获得良好的泛化能力，但是其判定精度比较差，会导致强化学习利用其奖励漏洞从而获得虚假的高回报。此外，这种二值化稀疏奖励会导致强化学习训练困难，并且训练过程不稳定。因此，如何获得稠密且可扩展的奖励函数是当前强化学习后训练的关键之一。其次，使用强化学习进行 VLA 模型后训练过程中会面临交互成本，因此提高强化学习样本利用率，减少后训练过程中的交互次数，是提高后训练效率的关键。从强化学习策略训练的角度来看，其中的关键问题在于如何获取稳定的价值函数，以及如何从离线数据阶段稳定地过渡到在线交互阶段。最后，安全性是在真实环境中进行强化学习后训练不可忽视的问题。然而，在强化学习过程中，探索作为策略提升的重要动力之一，通常会使机器人产生一些危险性动作从而导致机械臂与环境发生碰撞等问题。设计新的强化学习探索方法，同时兼顾样本效率和安全性，是实现 VLA 模型在真实场景中自主地持续迭代能力的前提。

## 7 VLA 模型评估

当 VLA 模型训练完成后，在真正部署应用与实际的生产生活前，需要对 VLA 模型进行全面的评估。合理的评估方法有利于评判不同 VLA 模型之间的性能差异，定位 VLA 模型问题，为 VLA 领域的下一步发展提供指导方向。本文首先对 VLA 评估指标进行了梳理，然后对目前 VLA 模型评估方法进行了梳理，将其划分为 3 类：基于真实环境评估，基于仿真器评估，基于世界模型评估。最后对目前 VLA 能力现状进行了总结。

### 7.1 泛化能力与评价指标

VLA 模型的泛化能力是评估模型能力的重要指标，通常会涉及到多个维度。本文将 VLA 操作泛化能力大致分为 3 个维度：形态泛化，任务泛化和环境泛化。其中形态泛化是指训练后的 VLA 模型部署到不同本体上，或本体发生变化的情况下，模型操作能力的变化[33-34,163,43,79]。这里的本体发生变化的情况有部署到与多种不同的机械臂上，或机械臂控制参数发生变化。任务泛化指的是与模型接收的语言任务指令相关物体发生变化的情况下，模型操作能力的变化。这类涉及到的情况会有任务语义变化、任务描述变化但语义不变、任务相关的物体种类变化、任务相关的物体颜色、位置或姿态发生变化[164-167]。环境泛化是指当机器人操作中的环境发生变化时，模型操作能力的变化[168,165]。涉及到的情况会有与任务无关的环境物体或布局变化、相机位置变化、光照变化、背景变化等。可靠的模型泛化性评估需要丰富且大量的实验验证，并通过统计评价指标来衡量模型在不同维度下的泛化能力。不同于前面衡量任务完成能力的绝对评价指标，RoboArena[169] 提出了一种能力对数的相对评价指标用于对比不同模型之间的性能差异，该指标考虑了模型的整体绝对

表 4: 不同 VLA 模型在真实环境下的测试结果[169]
Tab. 4: Performance of VLAs in Real-world[169].

| Rank | VLA 模型 | Score | SD | A/B Evals |
|---|---|---|---|---|
| 1 | $\pi_{0.5}$-DROID | 1883 | 26.1 | 339 |
| 2 | PaliGemma-FAST-specialist-DROID | 1851 | 25.8 | 741 |
| 3 | $\pi_0$-FAST-DROID | 1814 | 24.8 | 505 |
| 4 | PaliGemma-VQ-DROID | 1765 | 33.3 | 526 |
| 5 | PaliGemma-FAST-DROID | 1759 | 35.5 | 723 |
| 6 | PaliGemma-Diffusion-DROID | 1585 | 56.4 | 514 |
| 7 | DAM | 1213 | 210.5 | 53 |
| 8 | $\pi_0$-DROID | 894 | 28.5 | 781 |
| 9 | PaliGemma-Bining-DROID | 734 | 26.2 | 404 |

性能，任务对模型的难易以及模型对任务的擅长程度，通过双盲成对模型策略，大幅降低了人为主观因素对模型评价结果的影响。

目前，机器人操作任务成功率是评价 VLA 模型操作性能的主要指标。该指标可以直接评价 VLA 模型完成任务的能力，但是由于无法反映模型完成任务的过程，不利于分析失败场景下 VLA 模型的问题。因此，也有很多研究工作采用过程评估指标来分析 VLA 模型在完成操作任务过程中的表现。例如，对于需要多个操作步骤才能完成的任务，使用分阶段任务成功率来评估 VLA 模型的操作过程[170]。但这种方式只能提供稀疏过程反馈，并且无法用于评估单步骤的完成过程。INT-ACT[164] 使用意图来评估 VLA 模型在操作过程中的任务理解能力和推理能力。RobotArena∞[165] 使用任务进度实现更细粒度的模型在当前操作任务下的完成情况。此外，为了从更多的维度评估 VLA 模型的动作执行过程，RoboEval[170] 使用细粒度的行为指标，例如轨迹指标、空间指标和协调指标，来评估 VLA 动作执行过程中的安全性和平稳性。

### 7.2 基于真实环境评估

VLA 模型的长期目标是构建面向真实复杂任务的通用机器人策略，其训练过程使用了大量的真实世界数据。因此，在目前的很多 VLA 模型相关工作中使用物理机械臂在真实环境中对 VLA 模型性能进行测试。在 VLA 模型的评估过程中，通常会设置与训练数据分布相符的分布内任务和与训练数据分布差异较大的分布外任务，通过任务成功率指标，评估 VLA 完成任务的能力。分布内任务通常是指机器人系统设置，环境设置，以及任务设置与训练集中所包含的数据相符，用于评估 VLA 模型操作能力；分布外任务通常是指任务或环境设置发生一些变化，例如环境背景、光线、视角、操作对象种类等等，通常用于评估 VLA 模型的泛化能力。但由于操作物体种类选择的随机性和摆放位置的不同，导致不同模型之间很难进行公平的横向评估。针对此问题，功能性操作基准 (Functional Manipulation Benchmark, FMB)[181] 通过 66 种可 3D 打印标准化物体集、开源模仿学习框架及硬件规范，建立了可复现的评估体系。此外，系统性评估框架 Star-GEN 围绕视觉-语义-行为三模态定义 22 个泛化轴与 7 个任务类别[166]，解决了泛化能力定义模糊问题。目前真实环境中的 VLA 模型评估通常需要人设置环境并监督操作完成。一方面，人为主观因素可能会影响环境设置差异，从而导致 VLA 模型评估偏差；另一方面，人工操作的范式限制了这种评估体系很难大规模扩展。因此，AutoEval[182] 通过设计自动化场景重置和成功检测，取代人工操作，设计了一种全天候自主评估系统，提高了 VLA 模型评估系统的可复现性和标准化。

虽然在真实环境中评估 VLA 模型的性能最接近 VLA 部署后的性能，能够反映 VLA 在真实场景中的工作能力，但其面临的问题在于公平性和可复现性问题。由于不同的 VLA 模型预训练所使用的数据集以及传感器设置各不相同，例如所涵盖的机器人本体，相机的位置，相机的数量等等都有所差异。此外，由于预训练所使用数据集的差异性，导致真实环境测试时很难找到不同 VLA 预训练数据交叉下的实验场景进行布置。目前大部分 VLA 模型相关工作仅使用通过精心设计的，小范围实验场景和和测试条件对模型能力进行测试，很难反映真实的模型能力。RoboArena[169] 提出在 DROID[121] 数据集所覆盖的场景和机器人的基础上，通过众包的方式对不同的 VLA 模型进行了大量的双盲测试，并且使用模型能力对数这一相对评估指标而不是模型的绝对性能，可以更好地比较不同模型之间的性能以及更高的结果可信度。目前 RoboArena 的模型测试结果如表4中所示。其中 Score 表示能力对数，SD 是能力对数标准差，A/B Evals 表示双盲测试次数。能力对数值越高，表示模型能力越强。从表中的结果表明，目前 $\pi_{0.5}$ 是性能表现最好的模型，并且相比于均匀离散化和使用变分自编码离散化，FAST[80] 离散化动作空间综合表现更好。从该工作的实验测试结果中，我们可以发现，表中的 VLA 模型对不同视角、光照、背景具有鲁棒性，颜色/形状识别和分类能力较强，并且在直接物体操作任务上表现较好 (比如拾取放置、推动、简单开关)。但是对于工具使用和布料操作仍然比较困难，在需要精确对齐的任务（比如插入、堆叠等）上成功率低，并且多步骤推理和复杂语义理解能力有限。

### 7.3 基于仿真器评估

相比于真实环境评估，仿真器可以根据需求生成各种各样的场景，并且评估过程中不需要人参与，

表 5: 仿真器与 VLA 模型评估

Tab. 5: Benchmarks and Simulators for VLAs.

| 仿真环境 | 仿真引擎 | 输入 | 机器人 | 任务 | 相关方法 |
|---|---|---|---|---|---|
| CALVIN[171] | PyBullet | RGB/D，语言指令 | Franka Emika Panda | 长序列、语言指令驱动的桌面操作任务 | DreamVLA[69], GR-1[48], GR-2[49], RoboFlamingo[38], RoboVLMs[45], TriVLA[55], UniVLA[66], UP-VLA[64], VPP[47] |
| Franka-Kitchen[172] | MuJoCo | RGB/D | Franka Emika Panda | 厨房多物体交互，多目标组合任务 | HiRT[54] |
| SimplerEnv[173] | SAPIEN | RGB/D，语言指令 | WindowX, Google Robot | 多样化、语言指令驱动的桌面操作任务 | CogACT[41], HPT[34], Hume[61], LAPA[50], Octo[31], OpenVLA[40], RoboVLMs[45], RT-1[22], SpatialVLA[70], TraceVLA[127], UniVLA[66] |
| LIBERO[174] | MuJoCo | RGB/D，语言指令 | Franka Emika Panda | 专注于终身学习的程序化生成任务 | BitVLA[73], CoT-VLA[67], Fast ECoT[128], FLIP[175], Hume[61], NORA[129], OpenVLA[40], OpenVLA-OFT[78], SmolVLA[72], SpatialVLA[70], SP-VLA[176], TriVLA[55], UniAct[131], UniVLA[66], WorldVLA[68] |
| Meta-World[177] | MuJoCo | RGB/D | Sawyer | 50 种用于元学习/多任务学习的桌面操作任务 | HiRT[54], HPT[34], TinyVLA[178], VPP[47] |
| RLBench[179] | CoppeliaSir | RGB/D，语言指令 | Franka Emika Panda | 100 种大规模、带语言标注的多样化操作任务 | HybridVLA[100] |
| RoboMimic[180] | MuJoCo | RGB/D，语言指令 | Franka Emika Panda | 基于人类演示的模仿学习任务集 | HPT[34] |

表 6: 不同 VLA 模型在 LIBERO[174] 仿真场景下的测试结果

Tab. 6: Performance of VLAs in LIBERO[174].

| VLA 模型 | Spatial | Object | Goal | Long | Average |
|---|---|---|---|---|---|
| $\pi_0$[43] | **97%** | **99%** | 96% | 85% | 94% |
| $\pi_0$-FAST[43] | 96% | 97% | 89% | 60% | 86% |
| BitVLA[73] | **97%** | **99%** | 94% | 88% | 94% |
| CoT-VLA[67] | 88% | 92% | 88% | 69% | 84% |
| Fast ECoT[128] | 83% | 85% | 83% | 69% | 80% |
| GR00T N1[58] | 94% | 98% | 93% | 91% | 94% |
| Hume[61] | **97%** | **99%** | **99%** | **97%** | **98%** |
| NORA[129] | 86% | 88% | 77% | 45% | 74% |
| OpenVLA[40] | 85% | 88% | 79% | 54% | 77% |
| OpenVLA-OFT[78] | 96% | 98% | 96% | 91% | 95% |
| SmolVLA[72] | 93% | 94% | 91% | 77% | 89% |
| SpatialVLA[70] | 88% | 90% | 79% | 56% | 78% |
| SP-VLA[176] | 75% | 86% | 84% | 54% | 75% |
| TriVLA[55] | 91% | 94% | 90% | 73% | 87% |
| UniAct[131] | 77% | 87% | 77% | 70% | 78% |
| UniVLA[66] | **97%** | 97% | 96% | 92% | 95% |
| WorldVLA[68] | 73% | 88% | 80% | 27% | 67% |

具有更强的可复现性，评估结果也更客观，如表5所示。早期用于评估机械臂操作任务的仿真器有 RLBench[179]、Meta-World[177]、RoboMimic[183] 等，这些仿真器主要针对策略的控制能力进行评估，任务场景简单，并且不涉及语言指令。随着 VLA 模型的发展，基于仿真器的评估方法也在同步发展，例如 CALVIN[171] 和 LIBERO[174] 测试基准，在测试任务时加入了语言任务描述。CALVIN 不仅将语言指令加入到策略评估中，还引入了 RGB-D、本体感受和触觉等多模态信息。LIBERO 则是一个为了促进终身学习（Lifelong Learning）而提出的仿真器，其中包含 130 个不同的任务，涉及不同物体、不同空间关系和不同的任务目标。同时，LIBERO 还支持用户自行设计任务，为终身学习提供了有力支撑。CALVIN 和 LIBERO 引入了语言指令，是 VLA 模型评估的有效工具，但是这些仿真器渲染简单，轨迹分布呈现明显单峰分布，造成 VLA 模型评估结果与真实环境中的结果差异较大。SimplerEnv[173] 通过优化仿真器中的机械臂的比例-微分控制参数和运用视觉匹配（Visual Matching）技术，极大降低了仿真器与真机评估的差距。INT-ACT[184] 则在物体多样性、语言复杂性和场景复杂度上对 SimplerEnv 进行了进一步拓展，更适合对 VLA 模型的泛化性进行评估。VLATest[168] 使用视觉匹配增强的 Maniskill2[185] 仿真环境，对 7 个流行的 VLA 模型，从物体数量变化、光照变化、相机姿态和未见过物体操作等几个方面，全面评估了当前 VLA 模型操作能力。AGNOSTOS[167] 在 RLBench 仿真环境的基础上，通过操作对象与训练数据中的重叠程度，设置两种不同的未见过任务难度级别，详细评估了 VLA 模型跨任务泛化能力，发现现有的 VLA 模型在未见过任务上泛化能力严重不足。表6和表7分别展示了当前主流 VLA 模型在 LIBERO 和 SimplerEnv 环境上的上的测评结果。两者均涵盖了多种机械臂操作任务，例如抓取、放置、推物和打开容器等。为保证不同仿真器评估结果的可比性，本文采用任务成功率作为评价指标进行统计与展示，该指标用于衡量模型在给定语言指令下成功完成任务的比例，反映了模型在仿真环境中的整体操作能力与泛化水平。

### 7.4 基于世界模型评估

虽然仿真器评估具有低成本、可大规模扩展等优势，但是由于仿真器与真实环境的差异，导致其评估结果很难反映模型在真实环境中的性能。另一

表 7: 不同 VLA 模型在 SimplerEnv[173] 仿真场景下的测试结果
Tab. 7: Performance of VLAs in SimplerEnv[173].

| VLA 模型 | Google Robot | | | | WidowX Robot | | | |
|---|---|---|---|---|---|---|---|---|
| | Pick Coke Can | Move Near | Open/Close Drawer | Put Object in Drawer | Put Carrot in Plate | Put Spoon on Towel | Stack Cubes | Put Eggplant in Basket |
| $\pi_0$[43] | 73% | 65% | 38% | - | 0% | 29% | 17% | 63% |
| $\pi_0$-FAST[43] | 75% | 68% | 43% | 62% | 22% | 29% | **83%** | 48% |
| CogACT[41] | 91% | **85%** | **72%** | 51% | 51% | **72%** | 15% | 68% |
| HPT[171] | 60% | 24% | 56% | - | - | - | - | - |
| Hume[61] | **97%** | 80% | 59% | - | **67%** | 58% | 46% | 73% |
| LAPA[50] | - | - | - | - | 46% | 71% | 54% | 58% |
| Octo-Small[31] | - | - | - | - | 10% | 47% | 4% | 57% |
| Octo-Base[31] | 17% | 4% | 23% | - | 8% | 13% | 0% | 43% |
| OpenVLA[40] | 16% | 46% | 36% | - | 0% | 0% | 0% | 4% |
| RoboVLMs[45] | 77% | 62% | 43% | 24% | 21% | 46% | 4% | 79% |
| RT-1[22] | 3% | 5% | 14% | - | 4% | 0% | 0% | 0% |
| SpatialVLA[70] | 86% | 78% | 57% | **75%** | 25% | 17% | 29% | 43% |
| TraceVLA[127] | 44% | 55% | 44% | - | - | - | - | - |
| UniVLA[66] | - | - | - | - | 56% | 53% | 3% | **81%** |

方面，随着生成式模型的发展，使用世界模型进行 VLA 模型评估也逐渐开始崭露头角。相比于仿真器评估，世界模型通过真实数据构建，具有更逼真的图像渲染和更真实的动力学响应。1XWM[186] 提出了第一个能够预测人形机器人接触和全身操作的世界模型，实现了精确的动作控制能力，使得能够在相同的观测条件下比较不同策略的决策效果。该模型可以帮助 VLA 模型架构设计者快速评估不同模型架构设计，并从训练过程中选择最佳的模型检查点。Vid2World[187] 则对预训练的视频扩散模型进行了改造，使其能够根据机器人动作生成对应图像，并利用改造的世界模型对 RT-1 模型进行了评估。WorldEval[188] 使用 LoRA 对预训练的视频生成模型进行了微调，构建了一个世界模型，并将 VLA 模型解码前的动作表征输入到世界模型，这种动作表征方式使得 WorldEval 获得了更准确的动作响应能力。WorldEval 对 OpenVLA 和 $\pi_0$ 等 VLA 模型进行了评估，结果与真机上的评估结果呈现出强相关性。到目前为止，使用世界模型对 VLA 模型进行评估仍然处在起步阶段，其性能依赖与视频生成模型的性能，并且现阶段世界模型对于动作的响应能力比较有限，一定程度上限制了该模型用于 VLA 模型评估场景。

### 7.5 VLA 模型泛化性能现状

随着 VLA 模型结构、预训练数据规模、训练方法的发展，VLA 模型性能也取得了一定的进步。早期的 VLA 模型，对于操作对象的空间泛化能力比较差，例如 RT-1, RT-1-X, Octo, OpenVLA 等模型，当被操作物体位置、姿态发生变化，环境光照发生变化时，操作性能会大幅下降[168]。随着 VLA 模型的发展，一些新的 VLA 模型在上述场景下的泛化能力得到了大幅提升，例如 RoboVLM, SpatialVLA, CogACT, $\pi_0$, $\pi_{0.5}$ 等模型，尤其是引入空间信息的 SpatialVLA 模型对于操作对象的位置泛化能力表现优异。但是目前的 VLA 模型在环境泛化、任务泛化和形态泛化这 3 个维度仍然面临诸多的挑战。在环境泛化方面，目前大部分 VLA 模型操作性能对于相机视角和背景变化比较敏感[165]，但 $\pi_0$ 和 $\pi_{0.5}$ 展现出良好的泛化能力[169]。在任务泛化方面，大部分 VLA 模型对于操作物体的空间关系变化和形状、大小变化并不敏感，但是对于语义类变化非常敏感，例如颜色变化、语言指令变化以及使用新物体时，模型性能下降明显[166]。但是从意图或任务完成度这种更细粒度的指标观察，大部分基于 VLM 训练的 VLA 模型具有实施任务的倾向，但是这种语义倾向并没有转换为动作执行能力[164]。Magma 使用了视觉-语言联合训练，因此在动作意图和语言动作上表现更好。此外，目前的测评结果表明[165-167]，当前 VLA 模型对跨任务的泛化能力非常弱，这也是目前 VLA 领域亟需解决的问题之一。对形态泛化方面，目前缺乏比较系统的 VLA 模型形态泛化评估研究，一方面是由于目前的 VLA 模型形态泛化能力本身比较差，几乎很难使用相同的参数跨本体完成任务，另一方面，如何划分更细粒度的维度，评估模型在不同本体上的操作能力，也仍然是当前领域的开放性问题。

### 7.6 挑战与发展趋势

高效合理的 VLA 模型评估方法对于 VLA 模型的发展至关重要。目前基于真实世界的 VLA 评估方法大多还是依赖手工设计的场景进行评估，场景

表 8: 国内外产业界发布的 VLA 模型

Tab. 8: Industrial VLA Models Released Domestically and Internationally.

| | 公司 | 代表模型 | 应用场景 |
|---|---|---|---|
| 国外 | GOOGLE | RT-2[11], Gemini Robotics[60] | 通用桌面操作场景 |
| | NVIDIA | GR00T N1[58], GR00T N1.5 | 人形移动操作场景 |
| | Figure AI | Helix[57] | 家居场景/工业场景/物流场景 |
| | Physical Intelligence | $\pi_0$[43]， $\pi_{0.5}$[79] | 家居场景 |
| 国内 | 字节跳动 | RoboFlamingo[38], GR-1[48], GR-2[49], GR-3[139] | 通用桌面操作场景 |
| | 阿里巴巴 | RynnVLA-001[189] | 桌面操作 |
| | 美的 | DexVLA[190], ChatVLA[51], ChatVLA-2[95] | 通用桌面操作场景 |
| | 银河通用 | GraspVLA[115] | 桌面抓放场景 |
| | 智元 | G0-1[65] | 桌面操作场景 |
| | 星海图 | G0[191] | 移动操作场景 |
| | 星尘智能 | DuoCore-WB[192], ControlVLA[193] | 移动操作/家居场景 |
| | 自变量 | WALL-OSS[194] | 桌面操作场景 |
| | 灵初智能 | DexGraspVLA[56] | 桌面操作场景 |

的全面性和有效性往往不足。传统的评估体系需要人工操作员重置场景和成功判定，人工监督阶段可能会引入偏见，并且限制了评估的吞吐量和可扩展性。基于仿真器的评估虽然能够缓解上述问题，但是现阶段仿真器中机器人和环境与真实世界的差异导致其评估结果很难反映模型在真实世界中的性能。世界模型的发展为 VLA 模型评估提供了新的思路。一方面，世界模型通过数据驱动的方式，大大缩减了评估时环境与真实环境的差异，使其评估结果与在真实环境中接近；另一方面，世界模型可以通过大规模部署发挥可扩展优势。然而现阶段世界模型本身仍然面临诸多问题。首先，基于视频生成模型的世界模型无法建立准确的物理交互关系，从而有可能导致产生物理虚假现象。其次，世界模型对于动作的响应能力决定了其评估 VLA 模型的能力。到目前为止，现阶段世界模型仍然依赖训练数据中的动作分布，很难对分布外动作产生合理的响应。此外，如何控制世界模型产生复杂而且丰富的测试场景，从而全面的评估 VLA 模型，仍是一个需要探索的问题。

## 8 VLA 模型在产业界的发展

纵观 VLA 发展历程，产业界对于 VLA 的发展也至关重要。表8中列出了国内外产业界发布的 VLA 模型及其应用场景。谷歌在 2023 年首次提出 VLA 概念，并展现了该模型的通用操作能力，掀起了学术界和产业界对 VLA 研究热潮。英伟达也发布了面向人形机器人操作场景的 GR00T N1[58] 和 GR00T N1.5，展示了利用人类数据和仿真合成数据训练 VLA 模型的可能性。国外创业公司 Figure AI 面向家具场景发布了分层模型 Helix，展现出优秀的长程任务处理能力；物理智能公司发布 VLA 基座模型 $\pi_0$[43] 和 $\pi_{0.5}$[79] 受到领域的广泛关注，并且成为当前 VLA 领域使用最广泛的基座模型之一。

国内产业界对于 VLA 模型的关注度也非常高。字节跳动在 2023 年底就提出并开源了 RoboFlamingo[38]，作为 VLA 基准也一定程度上促进了领域的发展。随后发布的 GR-1[48], GR-2[49] 以及近期的 GR-3[139] 也得到了领域的广泛关注。美的集团在通用桌面操作场景也进行了一些列的探索工作，包括 DexVLA[190], ChatVLA[51] 和 ChatVLA-2[95] 等工作。阿里巴巴近期也发布了面向通用桌面操作场景的 RynnVLA-001[189] 模型。此外，国内初创公司在 VLA 领域也是百花齐放。银河通用推出了面向桌面抓放场景的 GraspVLA 模型，并展示了合成数据在 VLA 预训练过程中的重要作用。智元发布了面向通用桌面操作场景的 GO-1。自变量也发布了面向桌面操作场景的 WALL-OSS[194] 模型。星海图和星尘智能针对移动操作场景分别发布了 G0[191] 和 DuoCore-WB[192] 模型。

由于通过堆砌数据可以实现能力扩展，VLA 模型在产业界中的关注度比较高，并且对于其寄予厚望。但是目前 VLA 模型能力还面临诸多挑战，如对环境敏感，任务泛化能力比较差等。大部分的工作仍处在研究阶段，尚未形成规模化的落地验证。因此，从产业应用中找到合理的应用场景，先实现小范围内商业化应用，再逐步推广应用场景，是持续推动 VLA 模型良性发展的关键。

## 9 具身操作的 VLA 模型展望

### 9.1 泛化能力

尽管 VLA 模型的目标是构建面向复杂任务的通用机器人策略，但目前 VLA 模型的泛化能力仍然面临诸多挑战。首先，VLA 模型是一个基于视觉

和语言输入的机器人大模型，稳定的视觉泛化能力是实现处理复杂多样环境的前提。然而，目前 VLA 模型完成任务的成功率对于视觉变化比较敏感，包括对于背景变化、光照干扰、视角变化、物体位置等等。现有很多研究工作也关注到这个问题，并从跨域数据联合训练、大规模合成数据预训练以及多传感器融合的角度尝试去解决这个问题。另外一个泛化的角度是对机器人形态的泛化。VLA 模型预训练所需要的大规模数据很难由单一形态的机器人采集，通常会由多种不同形态机器人采集的数据汇聚而成。如何高效的利用不同机器人采集的数据，实现 VLA 模型跨形态泛化，是实现真正的机器人基础模型的必经之路。目前的研究方向集中在如何构建隐式共享空间，或设计合理的网络结构，重复利用并挖掘机器人本体信息的通用特征，但实现真正稳定的跨形态泛化仍然需要时间去探索。此外，跨任务泛化是 VLA 实现通用策略的关键之一，需要 VLA 模型对语言理解、观测理解和动作生成建立高度耦合且泛化的关联关系，实现对未训练过的对象和任务泛化。目前 VLA 相关工作中对于跨任务泛化研究相对较少，有研究工作发现当前 VLA 模型几乎不具备跨任务泛化能力[167]。

### 9.2 精细操作

目前的 VLA 模型在处理日常生活任务和复杂的长程任务中展现出了非凡的能力，但是在处理精细操作任务，尤其是需要密集接触的精细操作任务中，其成功率相对比较低。这里的主要原因可能有 2 个方向：一方面是对于这种精细操作任务，人类操作员采集的遥操作数据的动作一致性比较差，导致模型拟合难度比较高。这里的一致性差通常是由于不同的操作员习惯不同，或同一个操作员在不同次采集过程中的倾向不同导致的。对于一般作业任务，这种不一致性通常影响会比较小，但是对于精细操作任务，微小的动作变化通常会导致任务失败，因此会对这种不一致行为更敏感。目前有些工作从如何获得高质量数据的角度尝试解决该问题[153]，但尚处于早期阶段，如何构建更高效的高质量数据采集方案仍然是一个有待探索的问题。另一种潜在的解决思路是重复利用现有的不完美数据，通过离线强化学习的方式[151]，构建对于数据质量的评估方法，从而避免让模型在不一致的动作中产生均匀化现象。另一方面是由于目前的 VLA 模型缺乏其他模态的信息建立对环境更全面的感知，从而构建更合理的动作映射。例如，力或触觉传感器通常可以在视觉相似或遮挡情况下提供更多维度的信息以供模型决策。但是相关的研究仍处于比较早期的阶段[14,92,13]，如何将触觉与语言和动作对齐，不仅让模型能够理解语义中关于交互过程的描述，也能够让模型生成相应的动作，仍然是一个非常有价值且有待解决的问题。

### 9.3 实时推理

模型在机器人上推理实时性是 VLA 模型处理动态变化环境的前提。目前实时推理面临的挑战来源主要有 3 部分：首先，目前大部分 VLA 模型架构和权重继承自 VLM 模型，参数量和计算量相对比较大，在机器人上有限的计算资源很难部署并实现实时控制。相比于经典的小模型视觉端到端控制，目前的 VLA 大模型很难满足机器人高频的实时控制需求。其次，相比于云上平台的大型算力池，机器人上的算力非常有限，很难支持大模型在机器人本体上的实时计算。最后，模型的不合理部署会导致算力资源不能被充分的利用，如何根据模型特性，充分利用云端计算资源分布式部署 VLA 模型，也是实现机器人上 VLA 模型实时推理的关键之一。目前相关的研究工作尝试从两个方向去缓解 VLA 模型部署与实时控制之间的冲突。其中一个方向是在不改变 VLA 模型吞吐量的情况下，提高动作输出频率。其中一种做法是采用动作分块的机制[27]，通过预测未来多个时间步的动作，在模型推理过程中使用历史预测动作，从而缓解模型推理时间长导致的动作执行延迟，但是这种方式对于高动态变化环境无法及时作出响应[75]。此外，还有一种做法是采用异步分层机制[52-53,58-59]，上层是大模型推理任务与环境相关信息，下层是小模型负责动作高频生成。这种异步机制可以充分发挥云计算和端侧计算协同的优势，但其难点在于如何训练该分层模型，以及部署过程中如何保障通讯的实时性。另外一种做法是对 VLA 模型本身进行优化[72]、裁剪以及量化[73]，通过这种方式使得 VLA 模型可以直接部署到有限的机器人计算资源上，保障系统的稳定性。但是如何在优化模型参数的同时，保持 VLA 模型的泛化能力，是该方向面临的巨大挑战。

## 10 结束语

VLA 模型作为一个新兴的研究方法，受到学术界和工业界的广泛关注。本文从具身智能系统的角度，详细介绍了 VLA 模型在具身操作中的作用，及其发展过程，对 VLA 模型的模型架构、训练数据，预训练方法、后训练方法以及模型评估等方法进行了详细的介绍。最后根据该领域的发展现状分析了 VLA 模型在泛化能力，精细操作和实时推理这 3 个方面面临的挑战，以及在产业界的发展和未来可能的发展方向。VLA 模型发展为实现机器人通用策略探索了一条可能的道路，增强了具身智能系统与环

境交互的能力，使得具身智能机器人在工业、家庭、服务、物流等行业具有广阔的应用前景。VLA 模型目前还处在快速发展阶段，本文希望能够对该领域研究人员和相关从业者提供参考和方向指导。

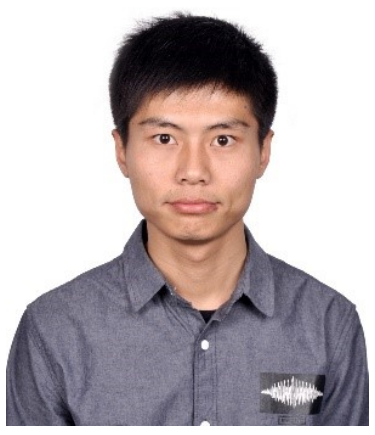

**李浩然**　中国科学院自动化研究所副研究员．中国科学院大学人工智能学院教师．2015 年获得中南大学学士学位．2020 年获得中国科学院自动化研究所控制理论与控制工程方向的硕士和博士学位．主要研究方向包括具身智能、强化学习及机器人学习．
E-mail: lihaoran2015@ia.ac.cn
(**LI Hao-Ran**　Associate Professor at Institute of Automation, Chinese Academy of Sciences, Beijing, China and also with the College of Artificial Intelligence, University of Chinese Academy of Sciences, Beijing, China. He received his B.S. degree from Central South University, Changsha, China, in 2015. He received his Ph.D. degree in control theory and control engineering at Institute of Automation, Chinese Academy of Sciences, Beijing, China in 2020. His research interests include embodied AI, reinforcement learning and robotic learning.)

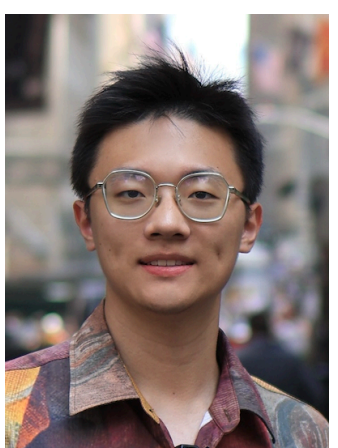

**陈宇辉**　中国科学院自动化研究所博士研究生．他于 2022 年获得北京理工大学学士学位和澳大利亚国立大学学士学位．主要研究方向包括具身智能、强化学习和机器人学习．
E-mail: chenyuhui2022@ia.ac.cn
(**CHEN Yu-Hui**　Ph.D. Candidate at the Institute of Automation, Chinese Academy of Sciences, Beijing, China. He received his B.S. degree from Beijing Institute of Technology, Beijing, China, and Australian National University, Canberra, Australia, in 2022. His research interests include embodied AI, reinforcement learning and robot learning.)

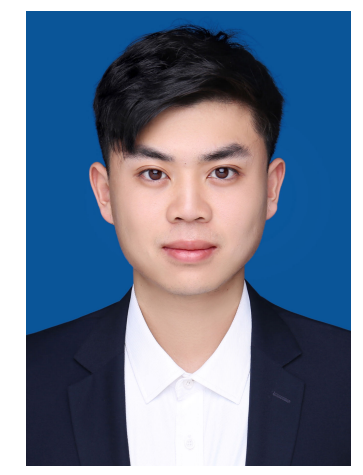

**崔文博** 中国科学院自动化研究所博士研究生．他于 2020 年获得东北农业大学学士学位他于 2023 年获得大连理工大学硕士学位．主要研究方向包括三维计算机视觉与具身智能．
E-mail: cuiwenbo2023@ia.ac.cn
(**CUI Wen-Bo** Ph.D. Candidate at the Institute of Automation, Chinese Academy of Sciences, Beijing, China. He received his B.S. degree from Northeast Agricultural University, Harbin, China, in 2020. He received his M.S. degree from Dalian University of Technology, Dalian, China, in 2023. His research interests include 3D computer vision and embodied AI.)

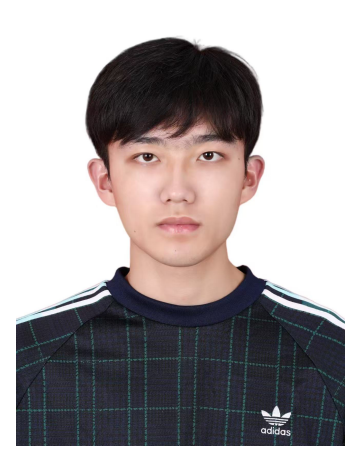

**刘卫恒** 中国科学院自动化研究所博士研究生．他于 2024 年获北京航空航天大学学士学位．主要研究方向包括具身智能、强化学习和 3D 视觉．
E-mail: weihliu2002@gmail.com
(**LIU Wei-Heng** Ph.D. candidate at the Institute of Automation, Chinese Academy of Sciences, Beijing, China. He received his B.E. degree from Beihang University, Beijing, China, in 2024. His research interests include embodied AI, reinforcement learning and 3D vision.)

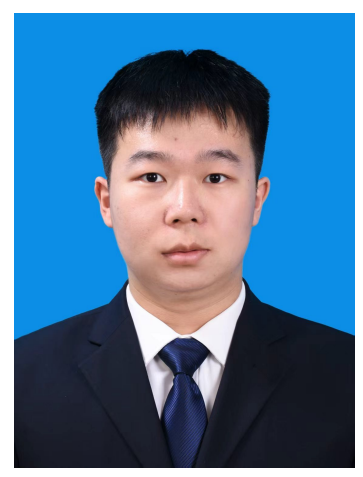

**刘锴** 中国科学院自动化研究所博士研究生．他于 2025 年获得西安交通大学学士学位．主要研究方向包括具身智能和世界模型．
E-mail: liukai2025@ia.ac.cn
(**LIU Kai** Ph.D. candidate at the Institute of Automation, Chinese Academy of Sciences, Beijing, China. He received his B.S. degree from Xi'an Jiaotong University, Xi'an, CHina, in 2025. His research interests include embodied AI and world models.)

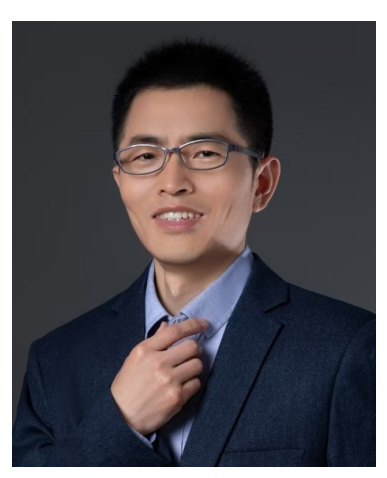

**周明才** 中国科学院自动化研究所副研究员．同时就职于北京中科慧灵机器人技术有限公司．他于 2010 年获得中国科学院自动化研究所博士学位．主要研究方向包括具身智能、计算机视觉、机器学习、计算机图形学、增强现实．
E-mail: zhengtao.zhang@ia.ac.cn
(**ZHOU Ming-Cai** Associate Research Fellow at the Institute of Automation, Chinese Academy of Sciences, Beijing, China. He is concurrently employed at Beijing Zhongke Huiling Robot Technology Co., Beijing, China. He received his Ph.D. from the Institute of Automation, Chinese Academy of Sciences, Beijing, China, in 2010. His research interests include embodied intelligence, computer vision, machine learning, computer graphics, augmented reality.)

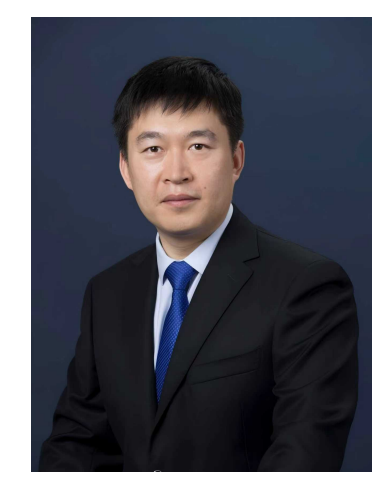

**张正涛** 北京中国科学院自动化研究所工业视觉与智能装备技术工程实验室教授．同时就职于北京中科慧灵机器人技术有限公司．他于 2004 年获得中国石油大学学士学位．他于 2007 年获得北京理工大学硕士学位．他于 2010 年获得中国科学院自动化研究所获得控制科学与工程方向博士学位．主要研究方向包括工业视觉检测和智能机器人．
E-mail: mingcai.zhou@ia.ac.cn
(**ZHANG Zheng-Tao** Professor at the CAS Engineering Laboratory for Industrial Vision and Intelligent Equipment Technology, Beijing, China. He is concurrently employed at Beijing Zhongke Huiling Robot Technology Co., Beijing, China. He received his B.S. degree from China University of Petroleum, Dongying, China, in 2004. He received his M.S. degree from Beijing Institute of Technology, Beijing, China, in 2007. He received his Ph.D. degree in control science and engineering from the Institute of Automation, Chinese Academy of Sciences, Beijing, China, in 2010. His research interests include industrial vision inspection, and intelligent robotics.)

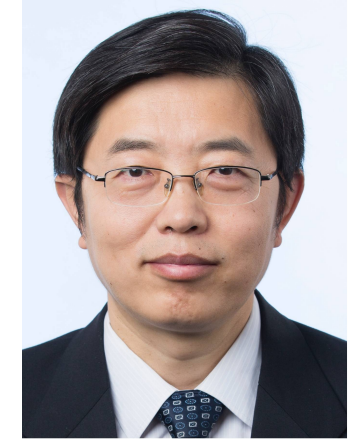

**赵冬斌** 中国科学院自动化研究所研究员．中国科学院大学教授．他分别于 1994 年、1996 年和 2000 年获得哈尔滨工业大学学士学位、硕士学位和博士学位．他于 2000 年至 2002 年在清华大学做博士后．主要研究方向为强化学习、具身智能、智能驾驶、智能博弈．本文通信作者．
E-mail: dongbin.zhao@ia.ac.cn
(**ZHAO Dong-Bin** Professor at the Institute of Automation, Chinese Academy of Sciences, Beijing, China, and the University of Chinese Academy of Sciences, Beijing, China. He received his B.S., M.S., Ph.D. degrees from Harbin Institute of Technology, Harbin, China, in 1994, 1996, and 2000 respectively. He was a postdoctoral fellow at Tsinghua University, Beijing, China, from 2000 to 2002. His research interests include reinforcement learning, embodied AI, autonomous driving and game AI. Corresponding author of this paper.)